\documentclass[referee,iicol]{sn-jnl}

\usepackage[utf8]{inputenc} 

\usepackage{graphicx} 
\usepackage{cancel} 

\usepackage{booktabs} 
\usepackage{array} 
\usepackage{paralist} 
\usepackage{verbatim} 
\usepackage{subfig} 
\usepackage{natbib}
\usepackage{soul}

\usepackage{graphicx}%
\usepackage{multirow}%
\usepackage{amsmath,amssymb,amsfonts}%
\usepackage{amsthm}%
\usepackage{mathrsfs}%
\usepackage[title]{appendix}%
\usepackage{xcolor}%
\usepackage{textcomp}%
\usepackage{manyfoot}%
\usepackage{booktabs}%
\usepackage{algorithm}%
\usepackage{algorithmicx}%
\usepackage{algpseudocode}%
\usepackage{listings}%

\newcommand{\bs}[1]{\boldsymbol#1}

\begin{document}

\title{Investigating Data Usage for Inductive Conformal Predictors}
\author[1]{\fnm{Yizirui} \sur{Fang}} \nomail
\author*[1,2]{\fnm{Anthony} \sur{Bellotti}}\email{anthony-graham.bellotti@nottingham.edu.cn}

\affil[1]{\orgdiv{School of Computer Science}, 
 \orgname{University of Nottingham Ningbo China},
 \orgaddress{\street{199 Taikang Dong Lu}, \city{Ningbo}, \postcode{315100}, \state{Zhejiang}, \country{China}}}
\affil[2]{ ORCID: 0000-0001-6317-5877}

\abstract{Inductive conformal predictors (ICPs) are algorithms that are able to generate prediction sets, instead of point predictions, which are valid at a user-defined confidence level, only assuming exchangeability. These algorithms are useful for reliable machine learning and are increasing in popularity. The ICP development process involves dividing development data into three parts: training, calibration and test. With access to limited or expensive development data, it is an open question regarding the most efficient way to divide the data. This study provides several experiments to explore this question and consider the case for allowing overlap of examples between training and calibration sets. Conclusions are drawn that will be of value to academics and practitioners planning to use ICPs.}


\maketitle



\section{Introduction}

Inductive conformal predictors (ICP) are machine learning algorithms that generate prediction sets, instead of point predictions, at a user-defined confidence level, and coverage, i.e. the probability that the true label is in the prediction set, is guaranteed to be at least this confidence level, only assuming that data are drawn from an exchangeable distribution. This is a property referred to as \textit{marginal validity} \citep{Vovk2005}. 
For example, a medical diagnosis system to predict one of 12 types of cancer could be built using ICP. For a new patient, instead of predicting just one cancer type it can predict a set of possible types (e.g. type A, D or F). If the medical doctor sets a confidence level of 90\% then ICP has a theoretical guarantee that the probability that the true cancer type is within this prediction set is 90\% or higher. 
The doctor is able to set a higher confidence level if she wishes and the ICP would meet this confidence level, but would need to output larger prediction sets to do so.
The validity property of ICP makes it valuable to achieve reliable machine learning, but we are also interested in achieving useful predictions. For ICP this means that prediction sets should be as small as possible. This is referred to as the \emph{predictive efficiency} of the ICP and can be optimized for different application problems.

ICP is built by partitioning a given development data set into a training, calibration and test data set. Typically, the partition is performed using random sampling with a predefined proportion of cases in each data set (e.g. split into proportions 50\%, 25\%, and 25\% respectively).
In many real world applications we are often confronted with the problem of limited data to use for constructing and validating machine learning models. In conventional machine learning, methods such as cross-validation and bootstrap testing can be used for efficient use of limited data \citep{Stone1974, Efron1997}.
The problem of limited data affects ICP, as well as other machine learning algorithms, and several solutions have been proposed to develop conformal predictors efficiently within different settings, such as cross-conformal predictor (CCP) \citep{vovk2015} as a form of cross-validation for ICP. However, there are limited results in the literature demonstrating how ICP actually behaves with different choices of the partitioning of the development data. 
\cite{Lofstrom2013} explore the use of data with Conformal Predictors, but focus on experimenting with different methods such as ICP along with proposed alternatives CCP, Bootstrap Conformal Predictors (BCP) and Out-of-Bag (OOB) test sets. They find evidence that CCP and BCP may not be valid, in the general case, and propose using OOB. Our study is different, in that we explore varying the size of data in training and calibration sets, with possible overlap, within the context of the ICP method only, since this is a popular version of conformal predictor with known marginal validity property.
We may suppose it is better to have more data as part of the training data than the calibration set, so that the model is trained sufficiently, but then if not enough data is left in the calibration set, this may effect the quality of the generated prediction sets and the empirical validity of results. It is an open question to ask how many is the right number in each set, and whether the amount of data required in each set changes with different confidence levels.

With very limited development data available, we may be tempted to share observations between training and calibration data sets, so that the partitions overlap, to enable more data to be available at both training and calibration stages. However, this would lead to a form of overfitting that we discuss in the methodology section and violates the assumptions that guarantee validity of ICP.
However, we may wonder how robust ICP's empirical validity is, with different degrees of overlap. We may speculate that just a small overlap of calibration and training data would have minimal effect on validity, as we observe it on test data, whilst giving more efficient prediction sets, but this remains an open question. We report results that explore these questions.

We give results of several experiments with different splits of the development data, based on different development scenarios. Three experiments are performed to explore the behaviour of ICP with different data partitions:
\begin{enumerate}[A.]
\item We suppose that the development set size is fixed and limited. A test set of fixed size is first drawn, then a training set with variable number of observations is then drawn. The remaining observations are then allocated to the calibration data set.
Thus, this experiment tests how ICP behaves with different partitions of data between training and calibration with no overlap between the partitions. 
\item In this experiment we consider different sizes of development data available. A test set of fixed size is first drawn, then equal size training and calibration data sets are drawn, but with a small fixed overlap of shared data.
This experiment tests the effects of training/calibration overlap, given different development data set size. 
\item This scenario considers the case of a limited fixed development data set, a fixed test set size and varying overlap between training and calibration set partitions from 0\% up to 100\%. This tests the case when available data is fully used but the overlap is varied.
\end{enumerate}
Each experiment requires a random split of the data into partitions. To account for the random perturbation, the experiments are run 200 times each and aggregate results are reported.
Each experiment are run across a range of confidence levels with an emphasis on high confidence levels since these are the ones that would be used in a real world setting. Confidence levels used are 0.2, 0.4, 0.6, 0.8, 0.9, 0.95, 0.975 and 0.99.
Since the experiments involve a large number of results, they are conducted on only one data set ``Covtype'' described in Section \ref{sec:Data}. This data set is selected since it is a publicly available multi-class problem which presents a sufficient challenge to ICP (see \cite{Bellotti2021}) that the data partition will give a perceivable effect.

ICP is typically set up as a wrapper algorithm for an underlying machine learning algorithm. In this study, the exact choice of the underlying algorithm is not particularly important, since we are focusing on the behaviour of the ICP. The artificial neural network (ANN) is used since this is a popular and topical algorithm in the machine learning community.

In the following sections, the methodology for this study is given including ICP and ANN, followed by experimental results and conclusions.

\section{Methodology}

\subsection{Inductive Conformal Predictors}\label{subsection:icp}
We set up the ICP framework following \cite{Vovk2005}.
\begin{itemize}
\item Let $\bs{z}_1, \cdots, \bs{z}_n$ be a sequence of $n$ examples $\bs{z}_j = (\bs{x}_j, y_j)$ from an exchangeable distribution, where $\bs{x}_j$ is a vector of $m$ predictor variables $\bs{x}_j \in \mathbb{R}^m$ and label $y_j \in \mathbb{Y}$ for some set of labels $\mathbb{Y}$.
\item Let $1$ to $k$ index the training data set and $k+1$ to $l$ index a calibration set, and $l+1$ to $n$ index a test data set, 
for $1<k<l<n$.
\item A conformity measure (CM) is any function 
$$A(\bs{z}) = \mathcal{A}(\bs{z}_1, \cdots, \bs{z}_k,  \bs{z})$$
such that $ \mathcal{A}$ is exchangeable: 
ie $ \mathcal{A}( \bs{z}_1, \cdots,  \bs{z}_k, \bs{z} ) 
=  \mathcal{A}(\bs{z}_{\pi(1)}, \cdots,  \bs{z}_{\pi(k)},  \bs{z} ) $ 
for all permutations $\pi$ of $1, \cdots, k$.
\item Let $\alpha_j = A(\bs{z}_j)$ denote the \emph{conformity score} for example $j$.
\item Let $\varepsilon$ be a preset significance level, so $1-\varepsilon$ is the confidence level for predictions.
\end{itemize}
The CM is intended to measure how typical an example $\bs{z}$ is, with respect to the training sequence.
We typically consider a CM to have the form $A(\bs{z}) = \mathcal{A}^p(\theta,  \bs{z} )$ for some vector of parameters $\theta = M(\bs{z}_1, \cdots,  \bs{z}_k)$
where $M$ is intended to be a model structure within which $\theta$ are estimated based on data $\bs{z}_1, \cdots,  \bs{z}_k$. For example, $M$ can be an ANN with $\theta$ being the vector of weights in the network.

The ICP is defined as the prediction algorithm that gives the prediction set at significance level $\varepsilon$ for a new example $\bs{x}$, based on the calibration set of examples $k+1$ to $l$ as
\begin{equation} \label{eq:predinterval1}
\Gamma^{\varepsilon}(\bs{x}) 
	= \left\{  
 \begin{array}{ll}
 \hat{y} \in  \mathbb{Y} : &
 \sum_{j=k+1}^l {\mathbb{I} \left[ A(\bs{x},\hat{y}) \ge \alpha_j \right]} + 1 \\
 & >  \varepsilon (l-k+1) 
  \end{array}
  \right\}
\end{equation}
where $\mathbb{I}$ is the indicator function. Notice that the prediction set is correct if it contains the true label $y$ corresponding to $\bs{x}$; i.e. $y \in \Gamma^{\varepsilon}(\bs{x})$. Assuming only that the calibration data set and any new test data sets are exchangeable, ICP predictions on the new data are guaranteed to be valid:
\begin{equation} \label{eq:validity}
\mathbb{P}(y_j \in \Gamma^{\varepsilon}(\bs{x}_j) ) \ge 1-\varepsilon
\end{equation}
for any new test example $j \in \{l+1, \cdots, n\}$ as is proved by \cite{Vovk2005}.
Indeed, for practical purposes, so long as the CM does not generate ties (i.e. examples with the same CM value), then exact validity is achieved:
\begin{equation} \label{eq:exact_validity}
\mathbb{P}(y_j \in \Gamma^{\varepsilon}(\bs{x}_j) ) = 1-\varepsilon.
\end{equation}
This is the setting we will use in this study.

When the label set $\mathbb{Y}$ is finite, this is ICP for classification. 
When $\mathbb{Y}=\mathbb{R}$, this is ICP for regression problems.
There are many different CMs that can be used, depending on the type of problem and underlying model $M$. For regression, the normalized non-conformity measure is popular  \citep{Papadopoulos2002InductiveRegression}.
In this study, we apply ICP in a classification setting and hence use a similarity metric between classifier prediction and outcome. We use CM of an example input $\bs{x}_j$ as the conditional probability given by the above-mentioned model $M$ given the sequence  $\bs{z}_1, \cdots, \bs{z}_k$,
\begin{equation}\label{eq:exact_conformity_measure}
    A(\bs{x}_j,y) = \mathcal{A}(\bs{z}_1, \cdots, \bs{z}_k,  (\bs{x}_j,y) ) = \mathbb{P}_M(y | \bs{x}_j)
\end{equation}
Hence, the conformity score for example $j$ in the calibration set is given by
\begin{equation}\label{eq:exact_conformity_score}
    \alpha_j = A(\bs{z}_j) = \mathbb{P}_M(y_j | \bs{x}_j)
\end{equation}
In this study, we allow a modification of this setting by allowing training and calibration sets to overlap, so in Equation (\ref{eq:predinterval1}), the index on the sum is changed to $j=c$ to $l$ for some $c \le k$.
The consequence of this is that the validity results in (\ref{eq:validity}) and (\ref{eq:exact_validity}) is no longer guaranteed. Indeed, this modification introduces a form of overfitting since the conformity score of an example in the overlap is dependent on training which uses itself hence introducing a bias. In such a case, we expect the distribution of conformity scores for overlapping examples to be calculated as higher than those when there is no overlap. This is demonstrated as an experimental result in Section \ref{exptC}.
Nevertheless, when developing ICP with limited data it is tempting to reuse data in this way, supposing that ICP is robust to some extent, perhaps for small overlap. In the experiments that follow, this speculation is tested.

\subsection{Artificial Neural Network}
\begin{figure}[ht]
    \centering
    \includegraphics[width=0.45\textwidth]{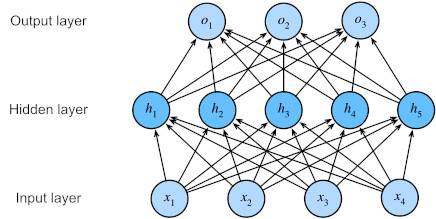}
    \caption{An example multilayer ANN with 1 hidden layer of $5$ neurons \citep{zhang2021dive}}
    \label{fig:example_mlpp}
\end{figure}

The underlying machine learning algorithm chosen for these experiments is the artificial neural network (ANN), due to its underlying flexibility and its popularity within the machine learning community.
The unit component of ANN is the neuron. Mathematically, this is a weighed sum of all input values with bias and a predefined activation function to emulate the biological neuron, i.e., it implements the function $y=\sigma\left({\bs{w} \cdot \bs{x} + b } \right) $, where $\bs{x}$ is the vector of inputs to the neuron, $\bs{w}$ is the weight vector, $b$ is a bias, $\sigma$ is the activation function, and $y$ is the output \citep{mitchell2007machine}. ANN is a collection of interconnected neurons, normally given as a series of layers from ANN inputs to outputs. An example is shown in Figure \ref{fig:example_mlpp}. This example has one hidden layer, but multiple hidden layers are possible.

For prediction, signals are propagated feed-forward from inputs through to outputs, via the neurons.
All the weights and biases are trained using the back-propagation algorithm \citep{mitchell2007machine}. This approach traverses the network from the output layer back to the input layer using gradient descent to optimize its training objective.

For multi-class classification tasks, the output layer of ANN is conventionally composed of $c$ neurons where $c$ is the number of labels in the set $\mathbb{Y}$. For this study, we use a probabilistic ANN, so output nodes output probabilities, and hence CM given in Equations (\ref{eq:exact_conformity_measure}) and (\ref{eq:exact_conformity_score}) can be used, and the CM parameters $\theta$ are all the weights and biases in the ANN.

\subsection{Performance metrics}

We consider measuring performance across repeated experiments $i=1$ to $200$.
For ICP, there are two types of performance measure. 

The first is \emph{coverage probability} which tells us the proportion of examples for which the true label is in the prediction set. On the test set this is
\begin{equation} \label{eq:coverage_prob}
c_i = \frac{1}{n} \sum_{j=l+1}^n {\mathbb{I} \left(y_j \in \Gamma^{\varepsilon}(\bs{x}_j) \right)}
\end{equation}
If the ICP is valid this should be as close to the confidence level as possible, therefore two measures are derived: 
$$b_i = c_i - (1-\varepsilon)$$
denotes the \emph{bias} between measured coverage and confidence level, and 
$$d_i = | b_i |$$
is the absolute \emph{diff}erence or error.
Hence, if ICP is valid, we expect $d_i \approx 0$ and the greater the value of $d_i$ the greater the deviation from validity (\ref{eq:exact_validity}). If the ICP is not valid, the bias $b_i$ measures the average direction of the error: if $b_i<0$ then the coverage is lower than required and expected by the validity result (\ref{eq:exact_validity}).

The second performance measure relates to the usefulness, or \emph{efficiency}, of the predictions. Typically, the smaller the prediction sets the more precise the prediction is. Hence a suitable measure of \emph{inefficiency} is the mean size, or width, of prediction sets on the test set:
\begin{equation} \label{eq:ineff_size}
w_i = \frac{1}{n} \sum_{j=l+1}^n {| \Gamma^{\varepsilon}(\bs{x}_j) |}
\end{equation}
The larger $w_i$ is, the less efficient the prediction sets are. Hence, the ICP performs best with smaller values of $w_i$. This measure corresponds to the N criterion given in \cite{Vovk2016}.

Since each experiment is run 200 times ($i=1$ to $200$), these measures are aggregated: mean values are given, along with empirical 95\% confidence intervals, assuming a normal distribution.
The aggregate of $d_i$ is called \emph{diff}, the aggregate of $b_i$ is \emph{bias} and the aggregate of $w_i$ is \emph{width}.

The potential for ICP to overfit when the calibration data set overlaps with the training set is explored in Experiment C, where the distributions of conformity scores of non-overlapping and overlapping data sets are provided.

\section{Experimental set-up and results}

The data set is described here, followed by details of experiments with results.

\subsection{Cover Type Data Set} \label{sec:Data}

This study is performed on the ``Covtype'' data set which is available from UCI Machine Learning repository \citep{Frank2010} and is chosen since the data set has sufficient examples for our study and the classification problem under ICP is not straightforward \citep{Bellotti2021} and so should show performance changes for our various experiments. 
Each example in the data is a small area (30m $\times$ 30m) of a forest represented by 12 attributes, including position, slope, elevation, lighting and soil type. Two of the attributes are coded as multiple binary variables thus giving a total of 54 predictor variables after one-hot encoding is applied. There are 7 class labels which are the primary major tree species growing in that area. 

There is high imbalance between the labels in this data set, but since we do not want to explore ICP under class imbalance for this particular study, and factor this out of these experiments, by randomly undersampling to ensure an
equal number of each class label. This gives 2747 examples for each class label, which is a total of 19229 examples. 

\subsection{Artificial Neural Network}
In this study, we design the ANN with multilayer perceptrons (MLP). We include one MLP with two hidden layers of $108$ and $12$ neurons, and $7$ neurons in the output layer representing the class labels of the prediction. The number of neurons in the first hidden layer is double the number of predictor variables. Stochastic gradient descent (SGD) is used as the optimizer, and the activation function is ReLU at hidden layers and sigmoid at the output layer to allow probabilistic predictions.

\subsection{Experiment A}

In experiment A, a randomly selected fixed test set of 5329 examples was first taken. This is a good size test set, whilst leaving 13900 examples for training and calibration sets which we found was a good number to achieve the best performance and reduce overfitting when training ANN, based on the learning curve. 
The remainder was split into training and calibration set with no overlap. Hence, the training set has size $k$ from 500 to 13400 and the calibration set has $13900-k$ examples.

Figure \ref{fig:expt_A}  shows results.
Generally they do not show deviation from validity (\emph{diff} is close to zero) and are unbiased, as we would expect. There is some suggestion of small downward bias around training set size = 6000 for confidence levels 0.2 and 0.9, but this is not clear since the upper bound on the 95\% confidence interval is very close to zero.

When the calibration data set becomes small ($<2000$), the bias becomes more unstable, with large confidence intervals, although on average, the ICP remains valid (\emph{diff} close to zero).

The right graphs (\emph{width}) are analogous to a learning curve. As the training data set becomes larger, predictive efficiency improves (i.e. $w_i$ gets smaller on average). Interestingly, however, the efficiency becomes unstable when the calibration set is too small ($<1000$), because the granularity of the $\alpha_i$'s becomes too low.

At low confidence level, predictive efficiency stabilizes at around 6000 to 8000 training observations, suggesting that is all that is required to achieve low confidence levels. However, for higher confidence levels, the efficiency converges in performance at about 12,000 training observations.

\subsection{Experiment B}

In experiment B, as with the previous experiment, a randomly selected fixed test set of 5329 examples was first taken. But then a training and calibration set of equal size was drawn randomly from 250 to 6500 examples, with an overlap of 250 examples between them. All other examples are discarded. 

Results are given in Figure \ref{fig:expt_B}. Since there is an overlap between training and calibration set, we would expect to observe some negative bias in the accuracy due to overfitting. However, this is generally not evident, since \emph{bias}=0 is typically within the 95\% confidence intervals. Only for combined sample size around 4000 (i.e. 2000 examples in training and calibration sets, with 250 overlap) for low confidence level ($<0.4$), there is some evidence of negative bias. This is because, on the one hand, with smaller training and calibration set sizes, model performance for high confidence levels is so poor that a significant difference in bias is not observable, and, on the other hand, with larger overall sample sizes, the relative size of the overlap is too small to impact the measurable deviation from validity.

Lower sizes of training and calibration sets leads to larger \emph{width}. This is to be expected since smaller training set sizes will lead to more inefficient prediction sets.
However, it is noticed that good results can be achieved at relatively small number of around 2000 examples.

\subsection{Experiment C} \label{exptC}

In experiment C, as with both previous experiments, a randomly selected fixed test set of 5329 examples was first taken. Then we consider a fixed number of 2000 examples to construct training and calibration set. This is deliberately small to simulate the case when the developer may be tempted to share examples between training and calibration sets, and to test how robust ICP is to such an action. Therefore, the procedure to construct training and calibration is to first randomly select $s$ examples, from $s=0$ to $2000$ which will be used in both training and calibration sets. The remaining $2000-s$ examples are randomly allocated to training and calibration sets in equal number. This means that training and calibration data will be the same size with $s+(2000-s)/2 = 1000+s/2$ examples each.
Notice that the two extremes, $s=0$ and $s=2000$ express no overlap and total overlap respectively.

Figure \ref{fig:expt_C} shows results.
Predictive efficiency improves quickly with overlap between training and calibration data sets. 
However, negative bias also emerges very quickly when there is an overlap, even when rather small ($250$ examples). Negative bias means that coverage probability on the test set is less than confidence level. This happens because the calibration set overfits to the ANN constructed using the training data, in terms of low conformity scores. This aligns with the discussion given in Section \ref{subsection:icp}. For the test set, which is independent of the training set, this will generate more efficient prediction sets but such that their coverage is not as high as would emerge from the overfitting calibration set. 
Curiously the bias is not evident until larger overlap sizes when confidence level is high; e.g. the bias becomes evident for overlaps of more than 750 examples when confidence level is 0.99. This suggests that ICP is more robust to training/calibration set overlap at higher confidence levels.

\begin{figure}[ht]
\centering
\includegraphics[width=0.5\textwidth]{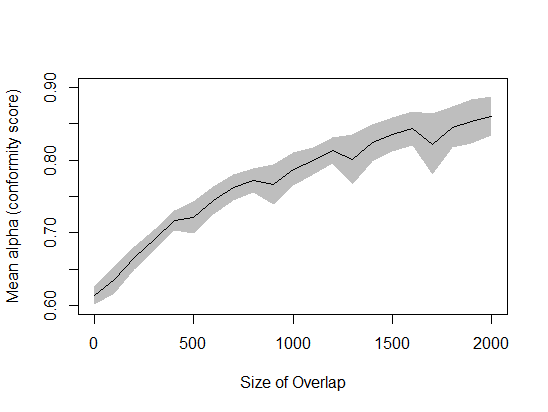}
\caption{Mean conformity score for each overlap size with 95\% confidence interval in Experiment C.}
\label{fig:alpha_dist_C}
\end{figure}

To explore the overfitting phenomenon, we compare conformity scores (alphas) for calibration sets with different size of overlap. For each run of Experiment C, the mean conformity score in the calibration set is computed, then the mean and standard deviation of mean conformity score across all 200 runs are computed and shown in Figure \ref{fig:alpha_dist_C} as mean with 95\% confidence interval ($\pm 2$ s.d.).
This graph shows that conformity scores in the calibration set become higher (i.e. increasing mean) with greater overlap. This is a natural consequence since overlapping examples in the calibration set will be \textit{conformal with themselves} as they occur in the training set. This indicates a type of overfit peculiar to ICP. Given the test set remains the same, it will lead to smaller prediction sets in the test set as a consequence of generally higher thresholds $\alpha_i$ for a label to appear in the prediction set as indicated in Equation (\ref{eq:predinterval1}).
On the other hand, this will then lead to lower coverage and invalid predictions; i.e. we expect that coverage probability will be less than confidence level.
This outcome is evident in the results presented in Figure \ref{fig:expt_C}.  

\section{Conclusion}

This study explores how data can be divided into training and calibration sets for ICP, when data is limited. In particular, the possibility of overlapping training and calibration sets was explored. The following two key conclusions are drawn from the experiments.
\begin{enumerate}
\item It may be possible to have a rather small calibration set (in our experiment, say, 2000 out of 13900) in order to maximize predictive efficiency, but it is important to be careful not to make calibration set too small, since this will lead to high variance of empirical validity and predictive efficiency, especially at higher confidence levels.
\item Our results suggests if the source data set is small, then introducing even a small amount of overlap between training and calibration data sets will lead to an invalid ICP and negative bias to validity. This lack of robustness suggests that developers should be careful when introducing an overlap, even if it yields improved predictive efficiency. With larger initial source data, a small overlap of training and calibration data will not give noticeable deviation from validity (bias), but then the overlap is too small to impact and improve the model in terms of predictive efficiency.
Alternatively, our results also suggest that ICP may be more robust to training/calibration data set overlap with higher confidence levels.
This is a useful insight for practitioners who wish to build reliable and safety critical applications with high confidence levels.
\end{enumerate}
These results will be of value to academics and practitioners planning to implement ICP with limited data.
For future work, similar experimental work can be conducted with ICP specifically in the context of imbalanced multi-class problems, and for regression problems.

\bmhead{Author Contributions}
Both authors contributed to the manuscript. The second author conceived the project, and the
first author did data analytics. Both authors wrote and reviewed the
manuscript.
\bmhead{Funding} There is no funding associated with this paper.
\bmhead{Conflict of interest} There is no conflict of interest.

\bibliographystyle{sn-basic}
\bibliography{CPdatause_text_v3}

\newpage
\onecolumn


\begin{figure}[ht]
\centering
\subfloat[Confidence level = 0.2]{\includegraphics[width=0.32\textwidth,height=0.26 \textwidth]{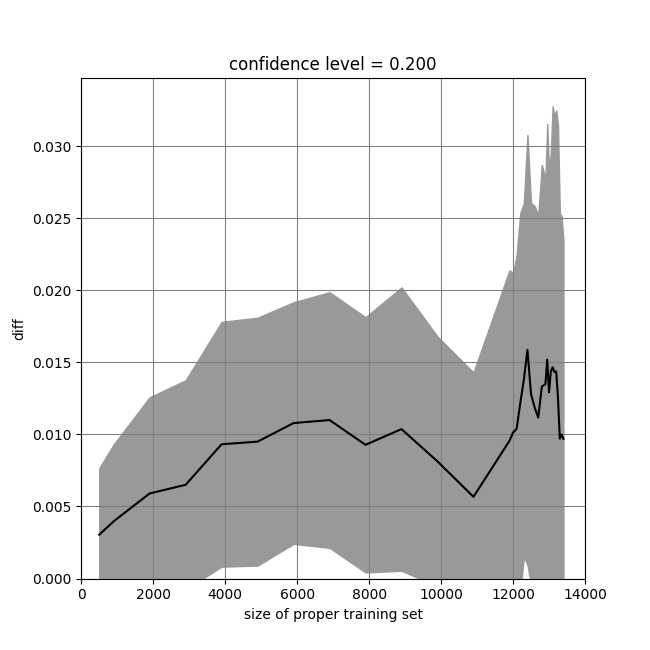} \includegraphics[width=0.32\textwidth,height=0.26\textwidth]{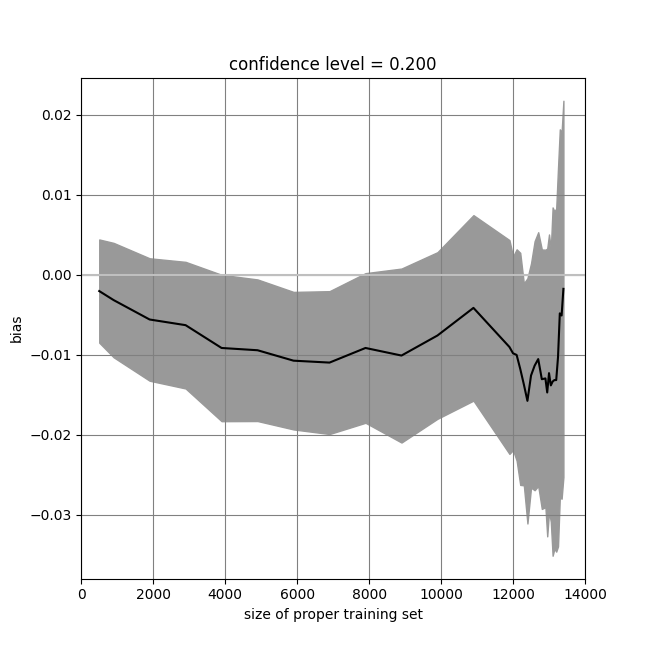}\includegraphics[width=0.32\textwidth,height=0.26\textwidth]{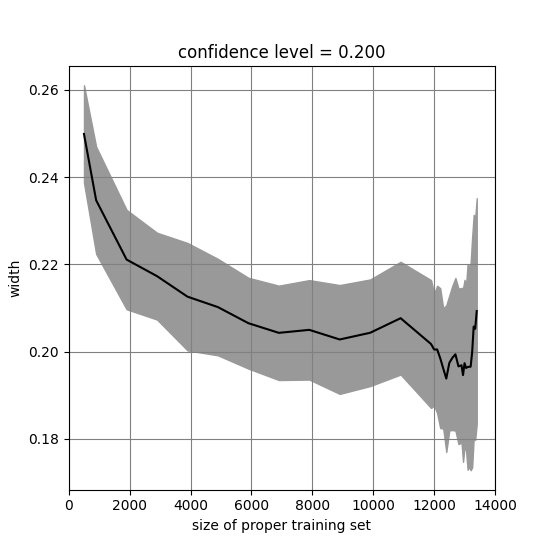}
\label{fig_no_overlap_index0}}
\hfill
\subfloat[Confidence level = 0.4]{\includegraphics[width=0.32\textwidth,height=0.26\textwidth]{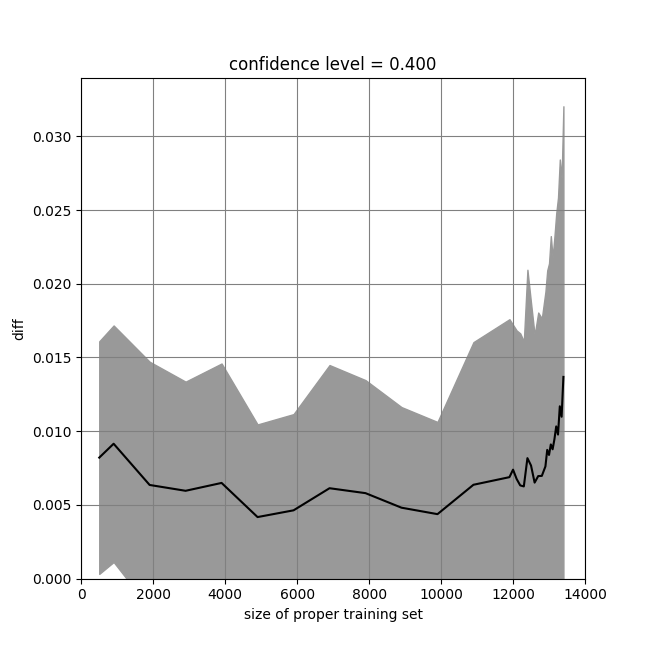}\includegraphics[width=0.32\textwidth,height=0.26\textwidth]{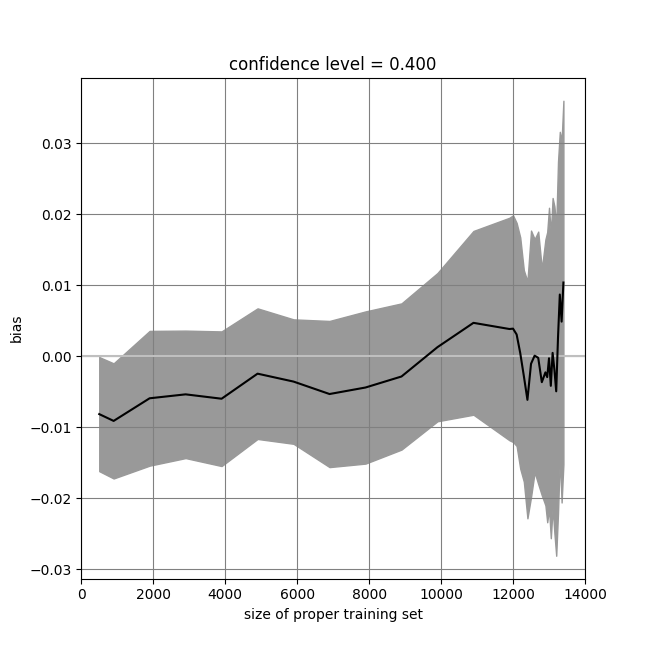}\includegraphics[width=0.32\textwidth,height=0.26\textwidth]{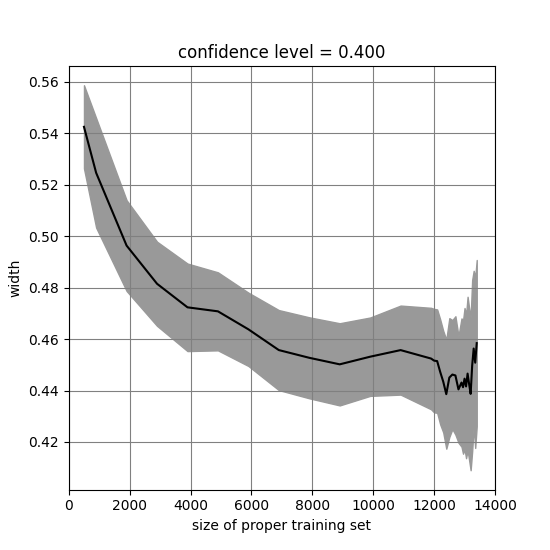}
\label{fig_no_overlap_index1}}

\subfloat[Confidence level = 0.6]{\includegraphics[width=0.32\textwidth,height=0.26\textwidth]{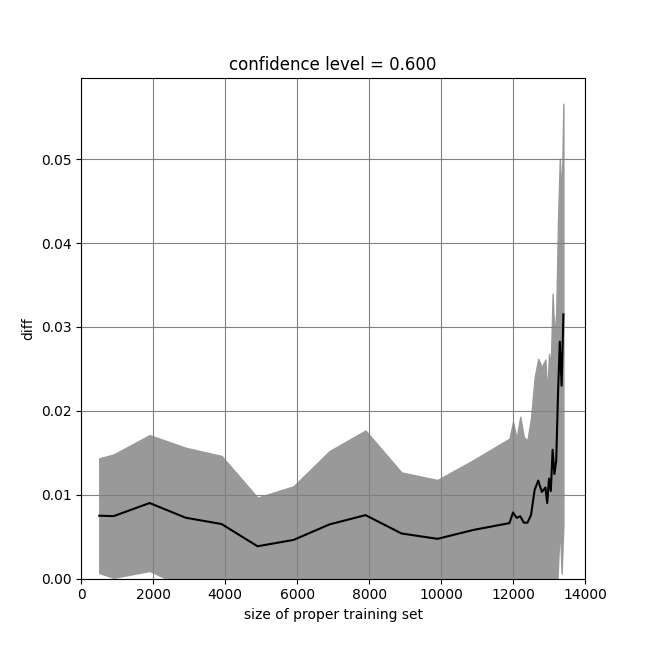}\includegraphics[width=0.32\textwidth,height=0.26\textwidth]{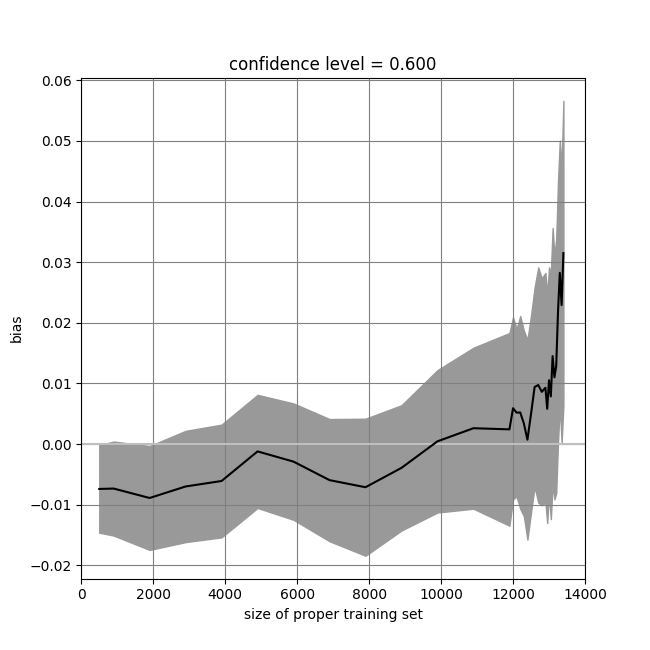}\includegraphics[width=0.32\textwidth,height=0.26\textwidth]{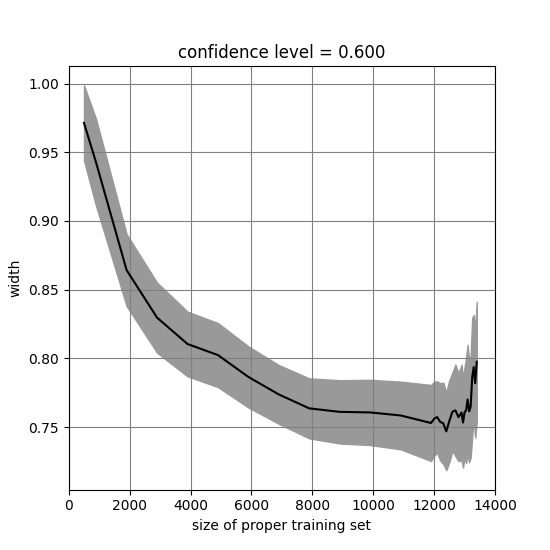}
\label{fig_no_overlap_diff_index2}}

\subfloat[Confidence level = 0.8]{\includegraphics[width=0.32\textwidth,height=0.26\textwidth]{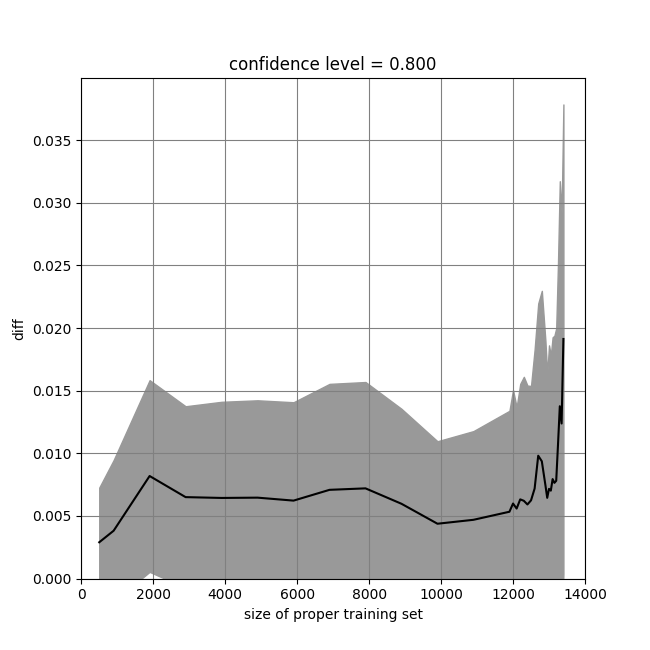}\includegraphics[width=0.32\textwidth,height=0.26\textwidth]{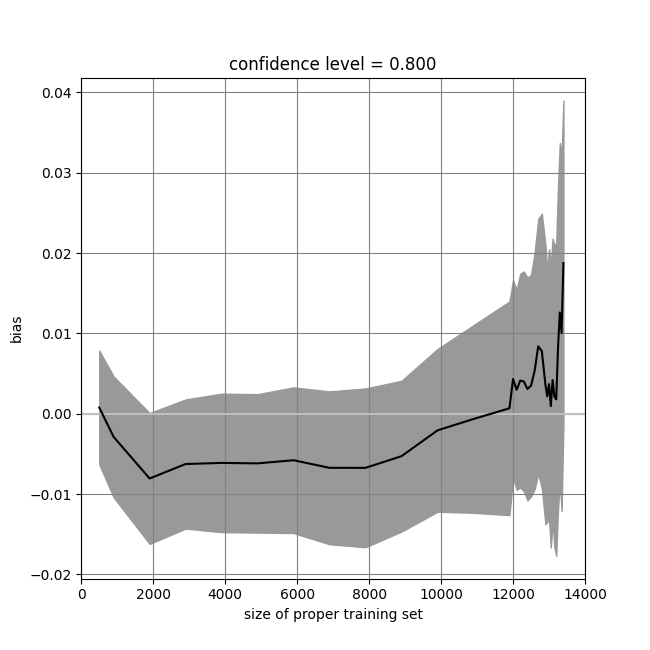}\includegraphics[width=0.32\textwidth,height=0.26\textwidth]{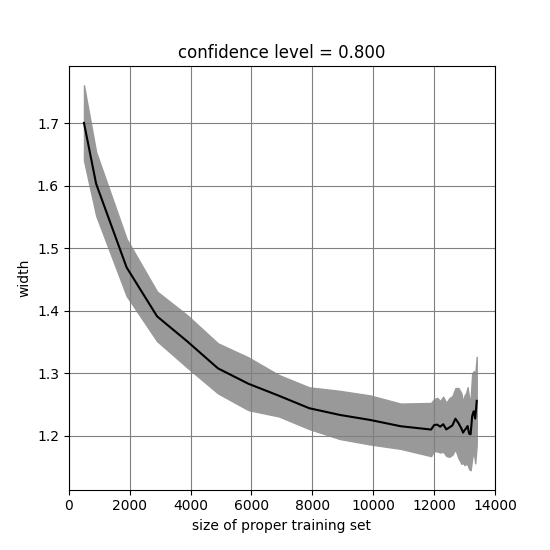}
\label{fig_no_overlap_index3}}
\caption{Experiment A: \emph{diff} (left), \emph{bias} (middle), \emph{width} (right) for different training set size (horizontal axes) and confidence levels, with 95\% confidence intervals.
\emph{Continued on next page.}}
\end{figure}
\begin{figure}[ht]
\ContinuedFloat
\subfloat[Confidence level = 0.9]{\includegraphics[width=0.32\textwidth,height=0.26\textwidth]{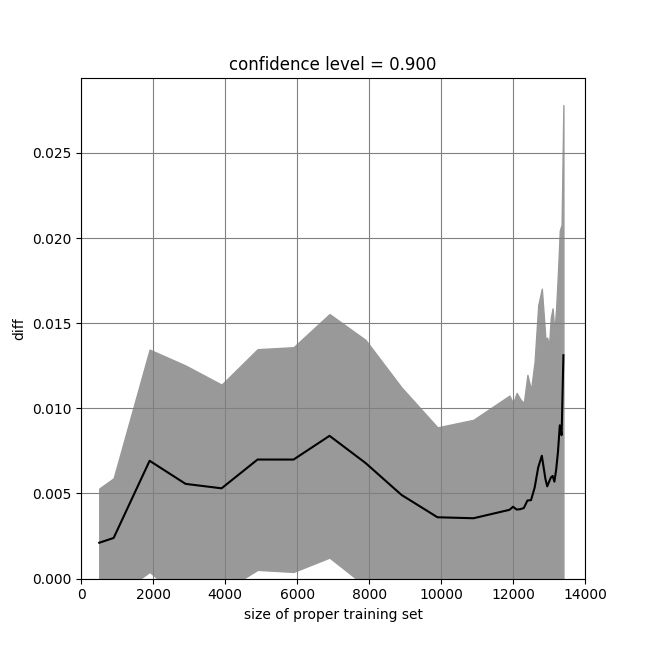}\includegraphics[width=0.32\textwidth,height=0.26\textwidth]{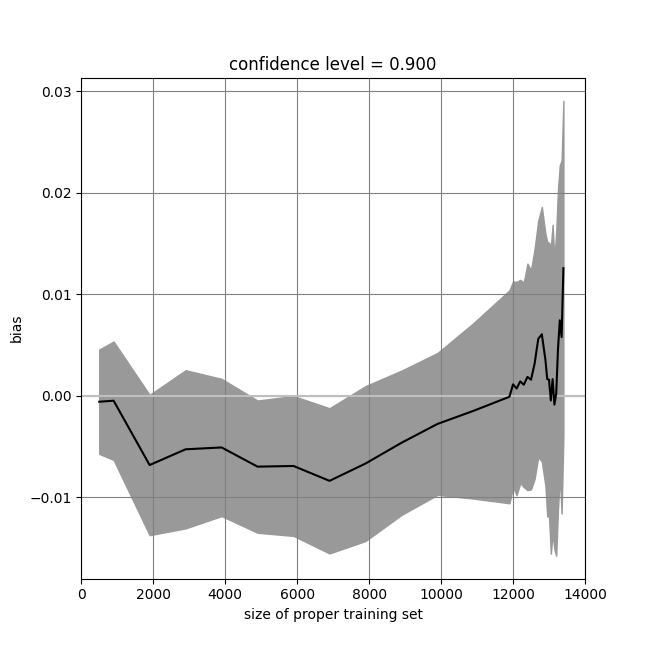}\includegraphics[width=0.32\textwidth,height=0.26\textwidth]{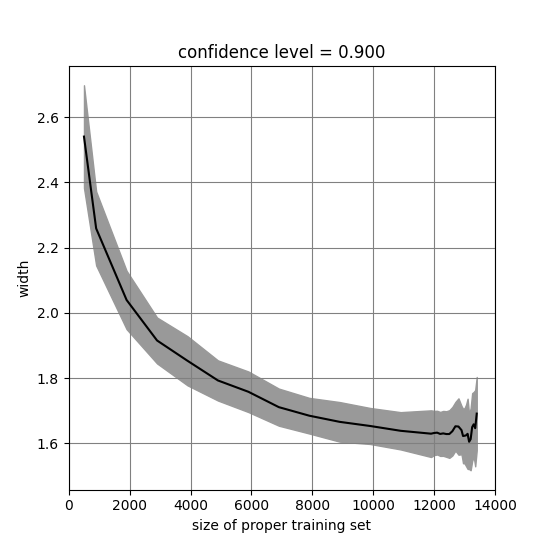}
\label{fig_no_overlap_index4}}
\hfill
\subfloat[Confidence level = 0.95]{\includegraphics[width=0.32\textwidth,height=0.25\textwidth]{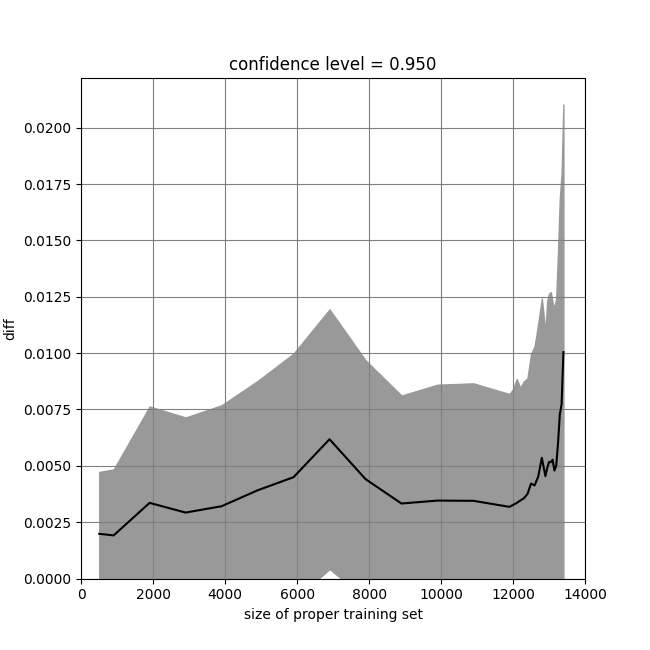}\includegraphics[width=0.32\textwidth,height=0.26\textwidth]{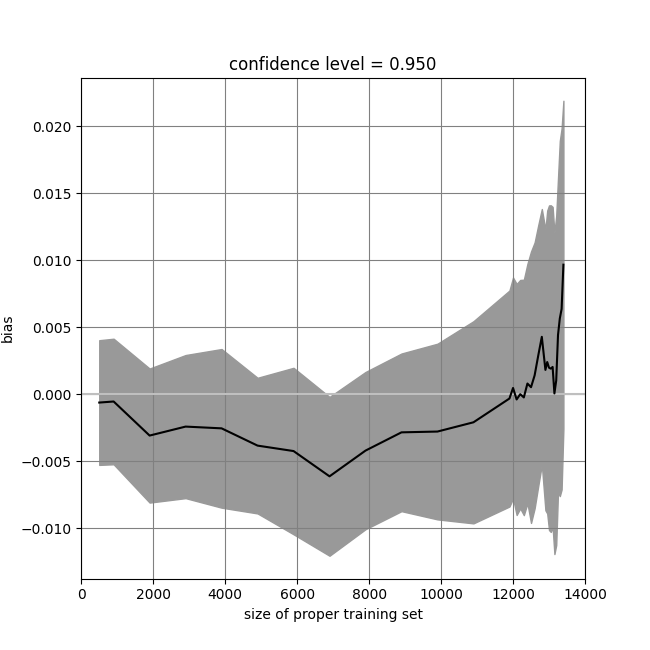}\includegraphics[width=0.32\textwidth,height=0.26\textwidth]{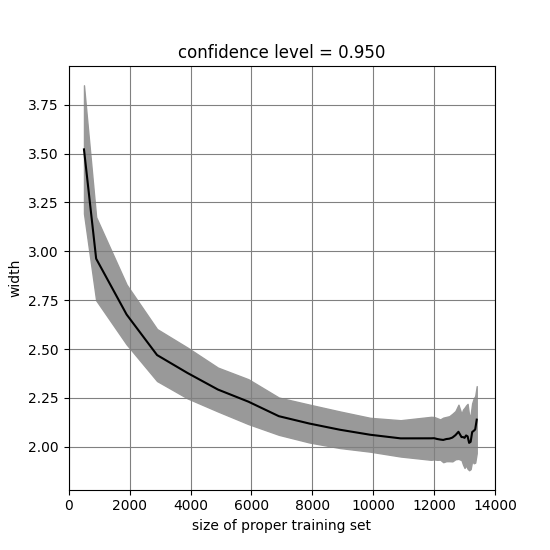}
\label{fig_no_overlap_index5}}
\hfill
\subfloat[Confidence level = 0.975]{\includegraphics[width=0.32\textwidth,height=0.26\textwidth]{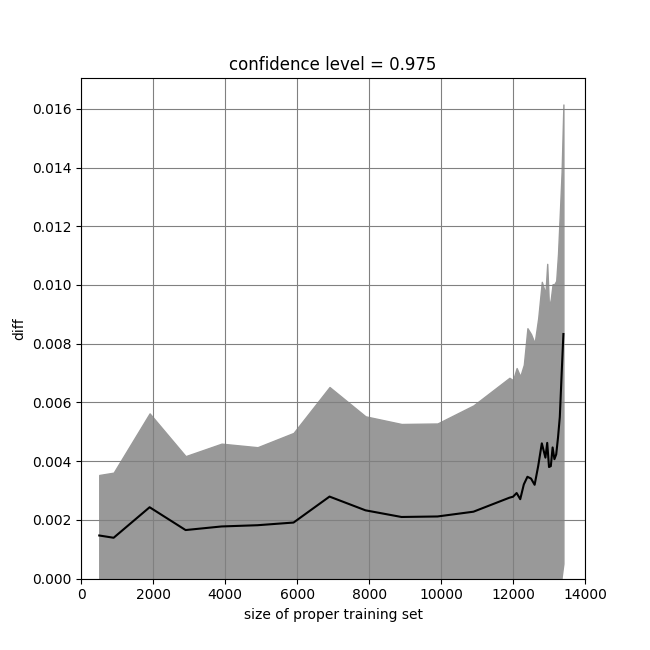}\includegraphics[width=0.32\textwidth,height=0.26\textwidth]{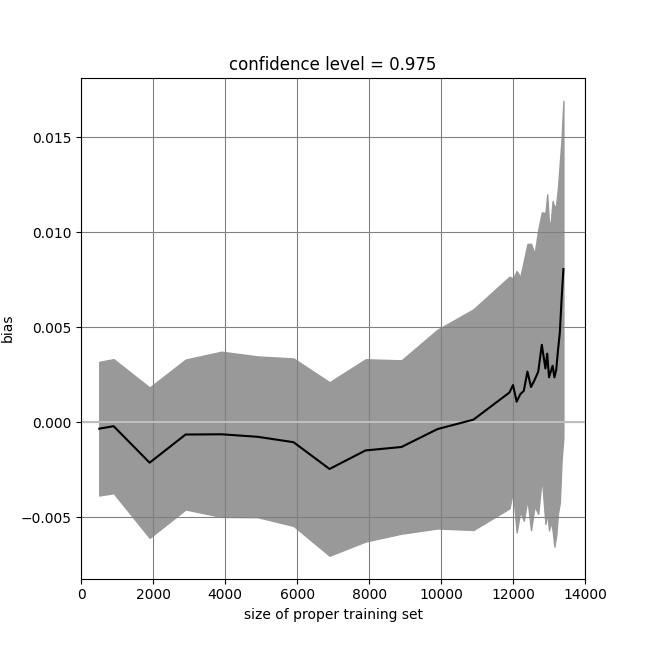}\includegraphics[width=0.32\textwidth,height=0.26\textwidth]{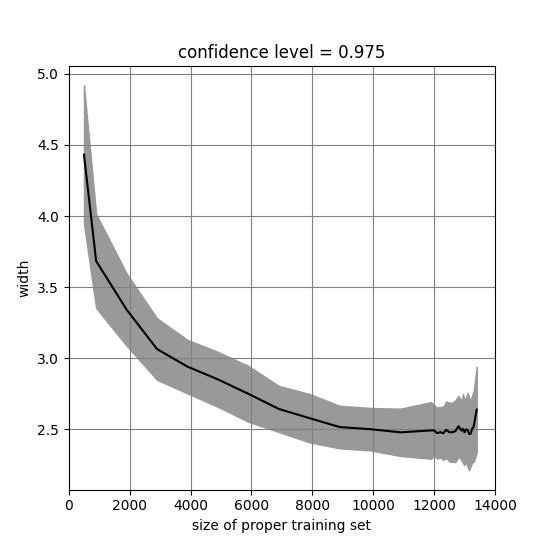}
\label{fig_no_overlap_index6}}
\hfill
\subfloat[Confidence level = 0.99]{\includegraphics[width=0.32\textwidth,height=0.26\textwidth]{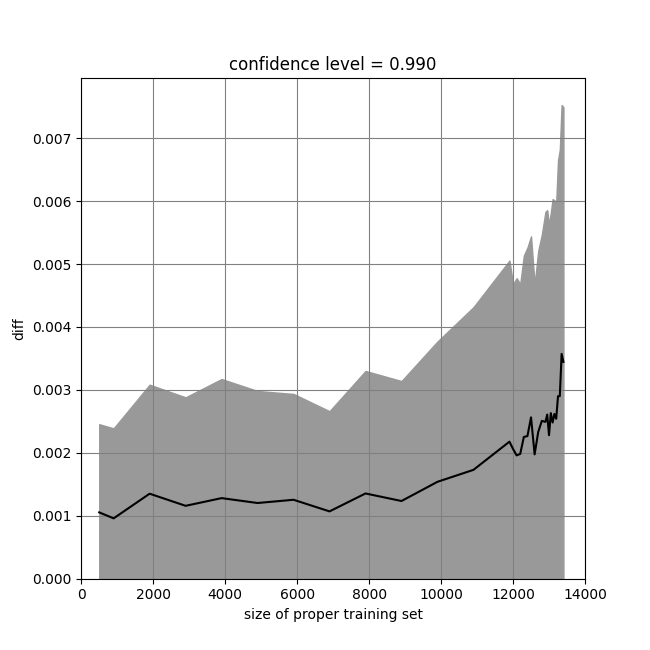}\includegraphics[width=0.32\textwidth,height=0.26\textwidth]{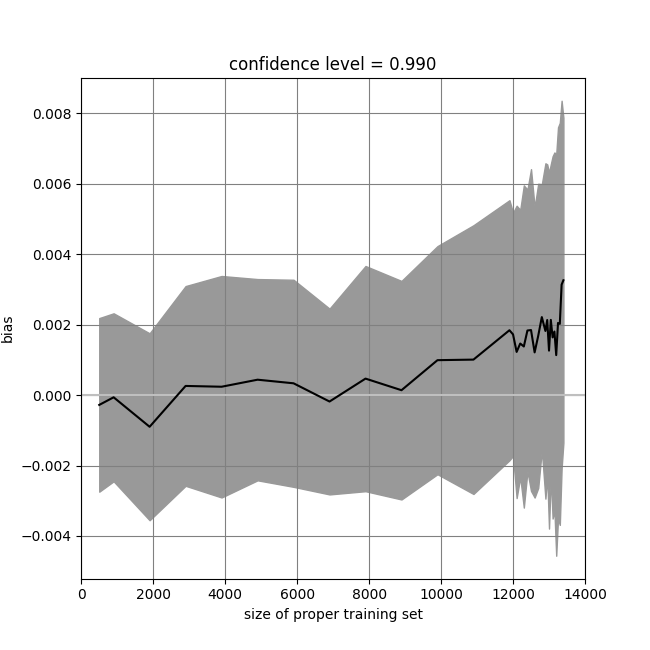}\includegraphics[width=0.32\textwidth,height=0.26\textwidth]{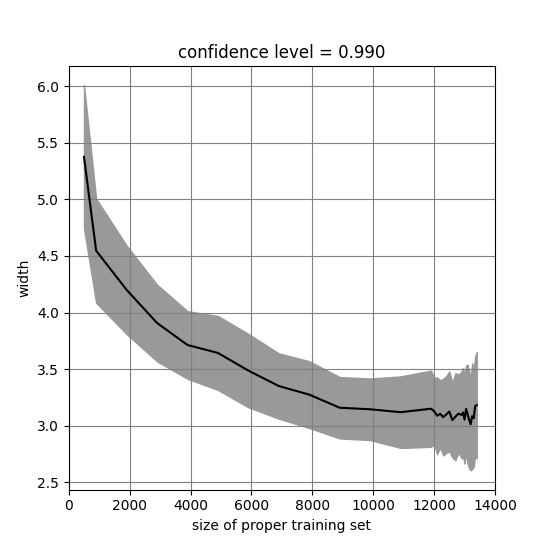}
\label{fig_no_overlap_index7}}
\hfill
\caption{Experiment A: \emph{diff} (left), \emph{bias} (middle), \emph{width} (right) for different training set size (horizontal axes) and confidence levels, with 95\% confidence intervals.}
\label{fig:expt_A}
\end{figure}

\begin{figure}[ht]
\centering
\subfloat[Confidence level = 0.2]{\includegraphics[width=0.32\textwidth,height=0.3\textwidth]{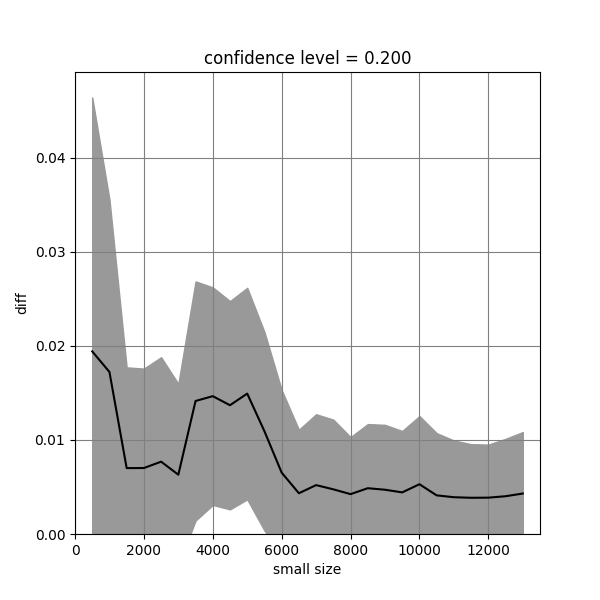}\includegraphics[width=0.32\textwidth,height=0.3\textwidth]{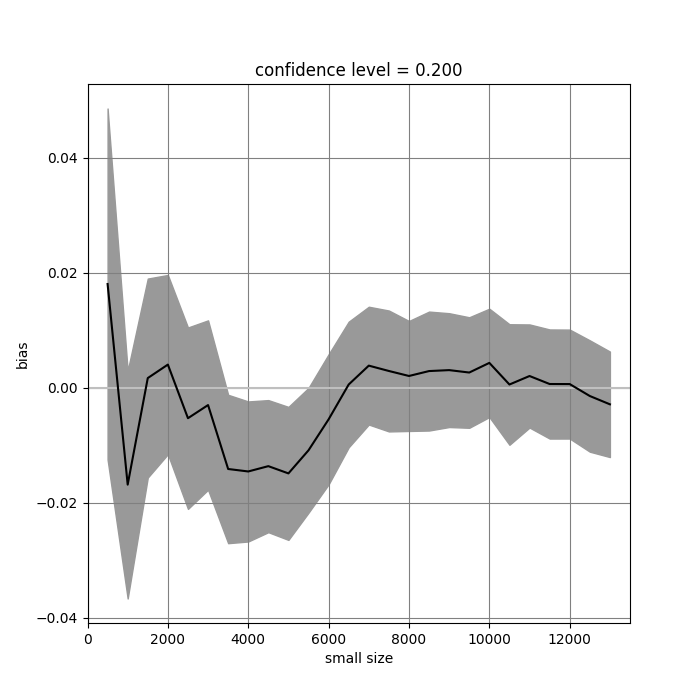}\includegraphics[width=0.32\textwidth,height=0.3\textwidth]{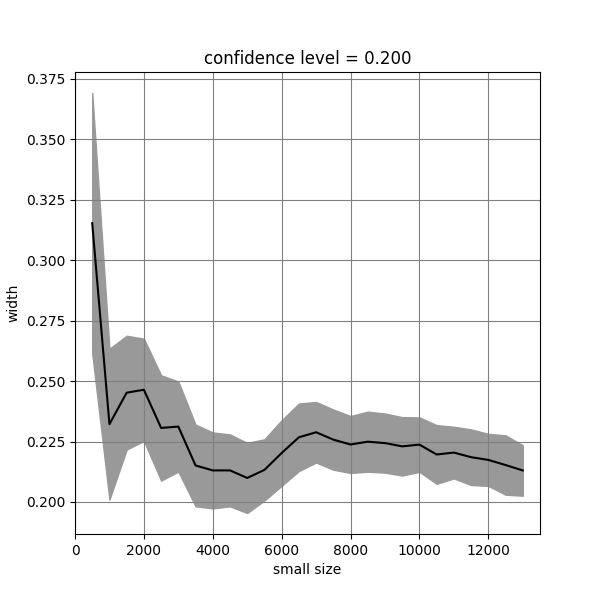}
\label{fig_small_size_index0}}
\hfill
\subfloat[Confidence level = 0.4]{\includegraphics[width=0.32\textwidth,height=0.3\textwidth]{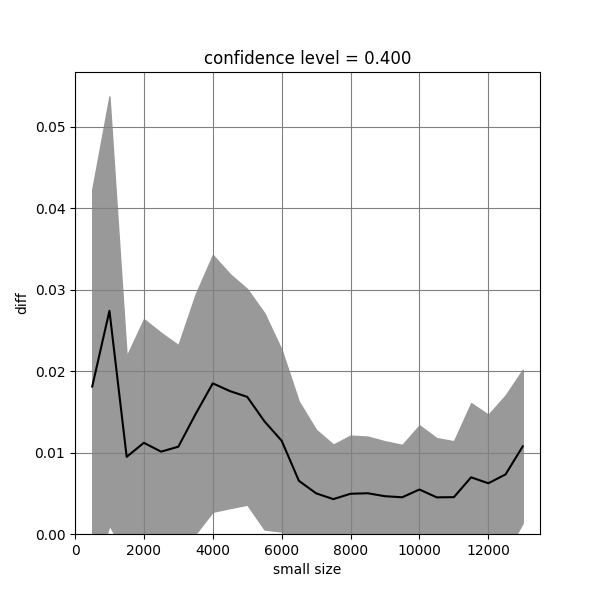}\includegraphics[width=0.32\textwidth,height=0.3\textwidth]{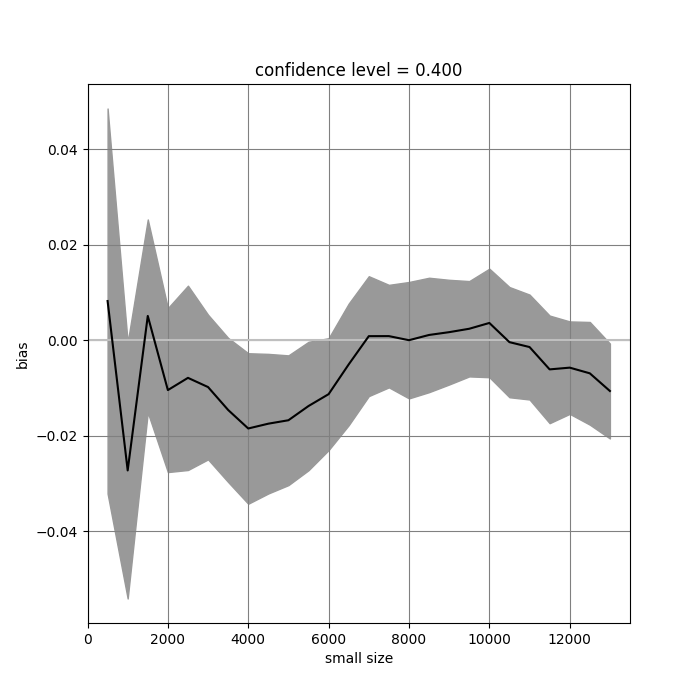}\includegraphics[width=0.32\textwidth,height=0.3\textwidth]{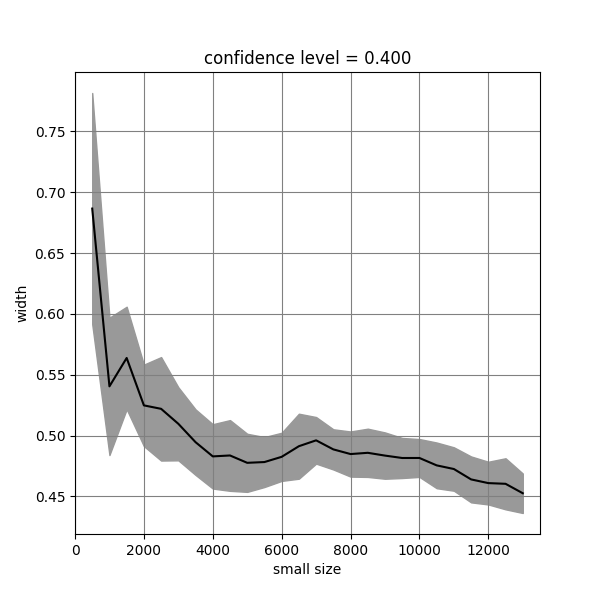}
\label{fig_small_size_index1}}
\hfill
\subfloat[Confidence level = 0.6]{\includegraphics[width=0.32\textwidth,height=0.26\textwidth]{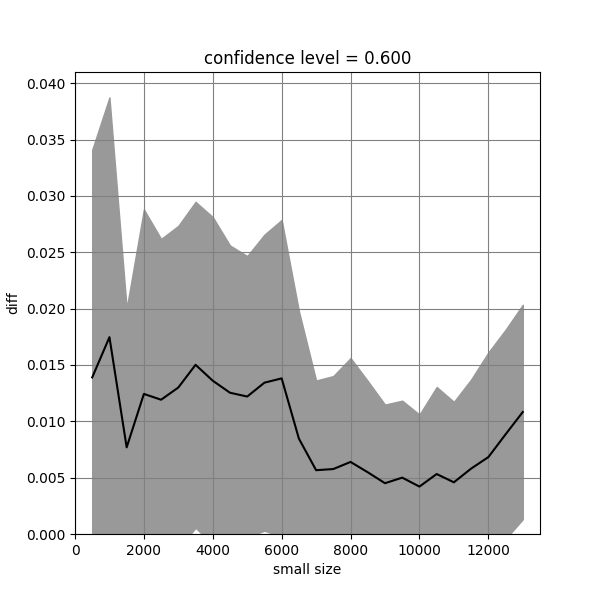}\includegraphics[width=0.32\textwidth,height=0.26\textwidth]{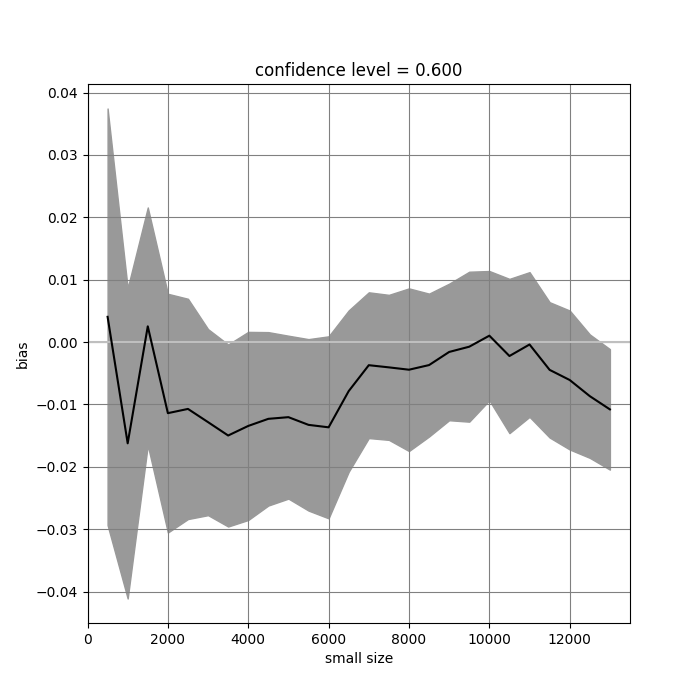}\includegraphics[width=0.32\textwidth,height=0.26\textwidth]{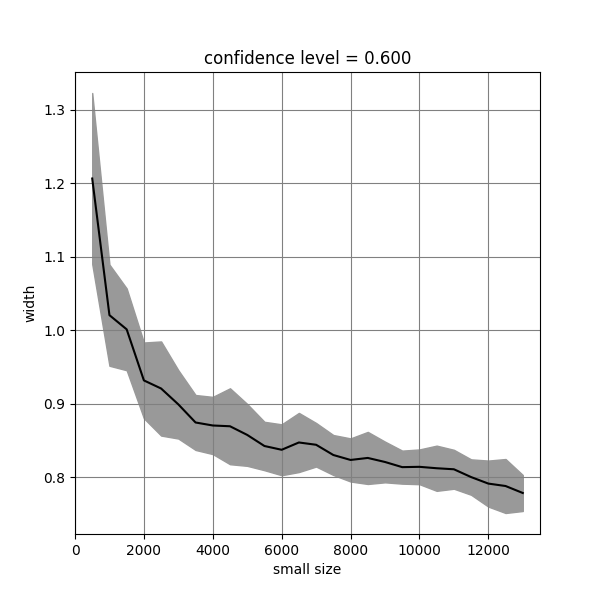}
\label{fig_small_size_index2}}
\hfill
\subfloat[Confidence level = 0.8]{\includegraphics[width=0.32\textwidth,height=0.26\textwidth]{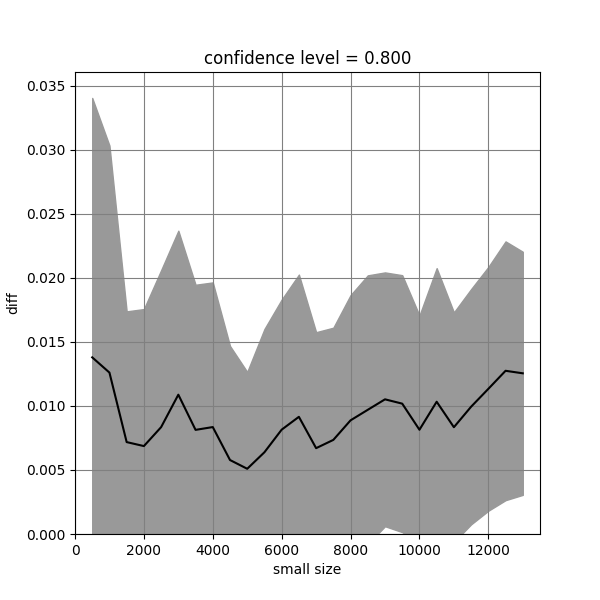}\includegraphics[width=0.32\textwidth,height=0.26\textwidth]{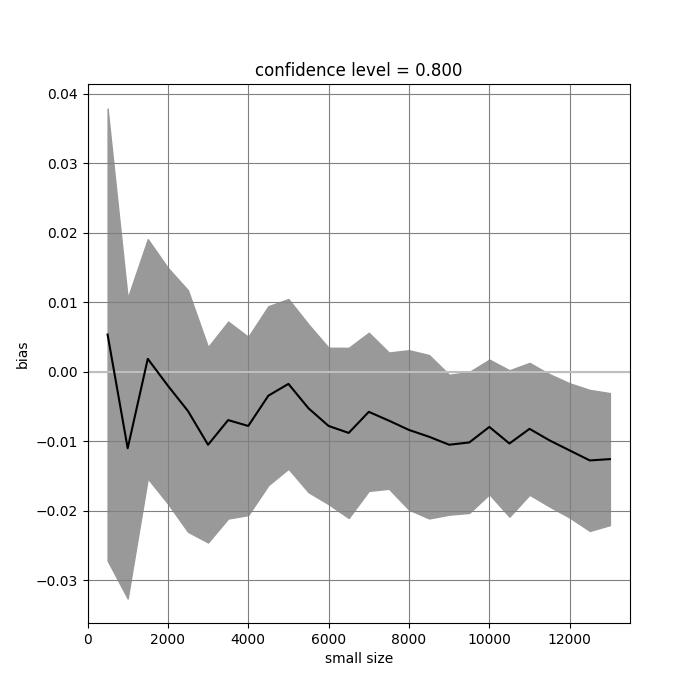}\includegraphics[width=0.32\textwidth,height=0.26\textwidth]{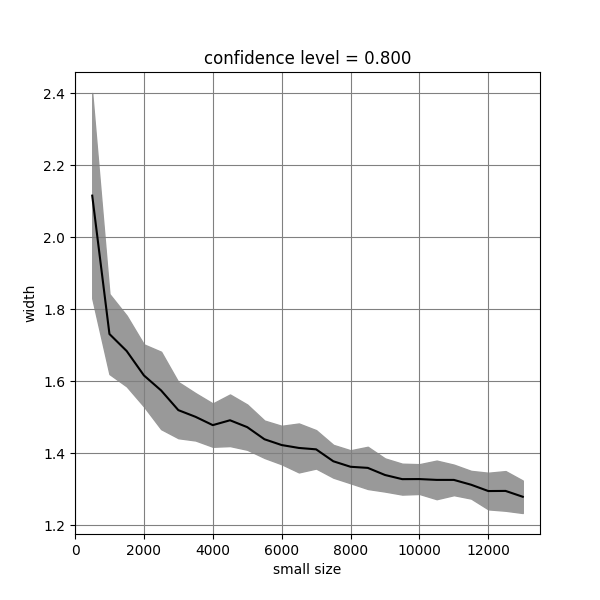}
\label{fig_small_size_index3}}
\hfill
\caption{Experiment B: \emph{diff} (left), \emph{bias} (middle), \emph{width} (right) for different combined calibration and training set size (horizontal axes) and confidence levels, with 95\% confidence intervals.
\emph{Continued on next page.}}
\end{figure}
\begin{figure}[ht]
\ContinuedFloat
\subfloat[Confidence level = 0.9]{\includegraphics[width=0.32\textwidth,height=0.26\textwidth]{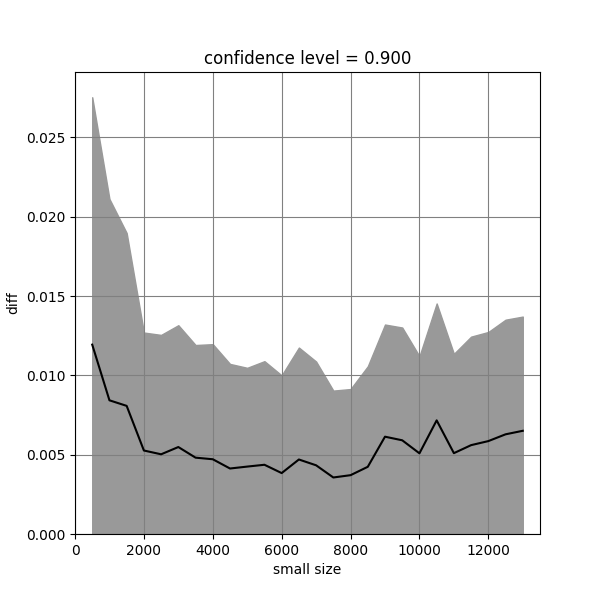}\includegraphics[width=0.32\textwidth,height=0.26\textwidth]{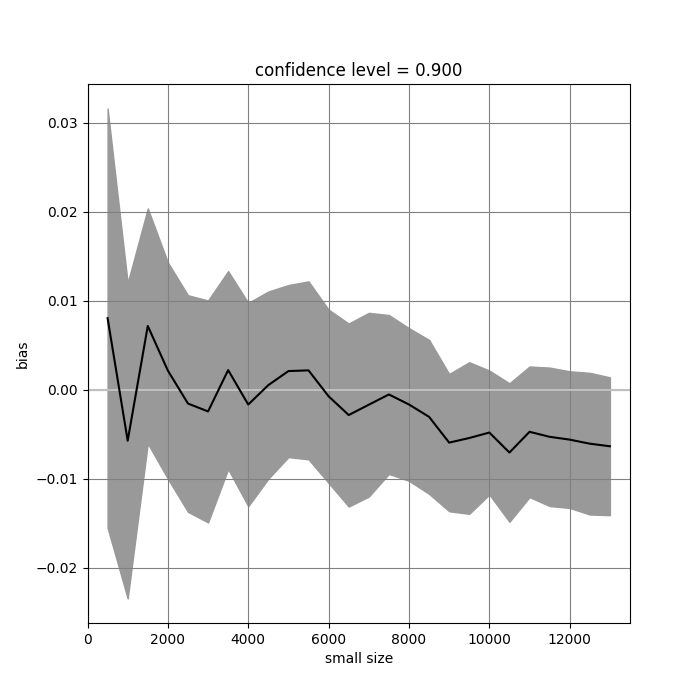}\includegraphics[width=0.32\textwidth,height=0.26\textwidth]{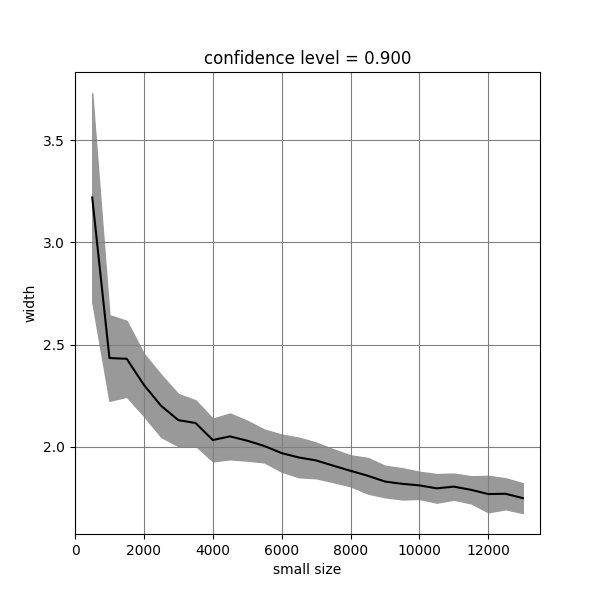}
\label{fig_small_size_index4}}
\hfill
\subfloat[Confidence level = 0.95]{\includegraphics[width=0.32\textwidth,height=0.26\textwidth]{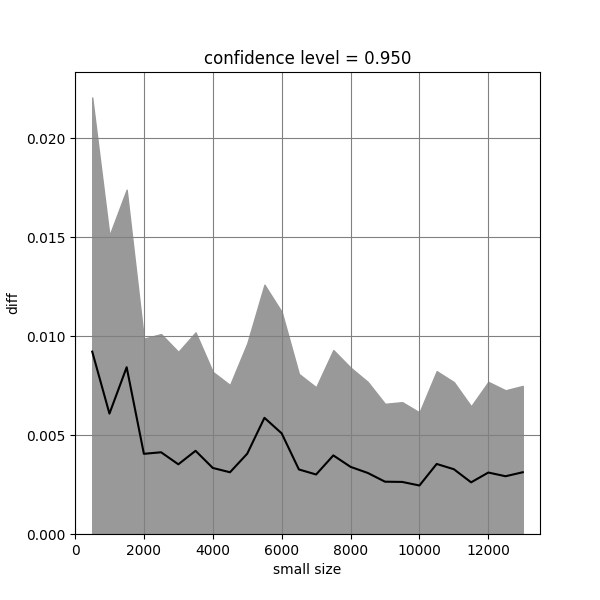}\includegraphics[width=0.32\textwidth,height=0.26\textwidth]{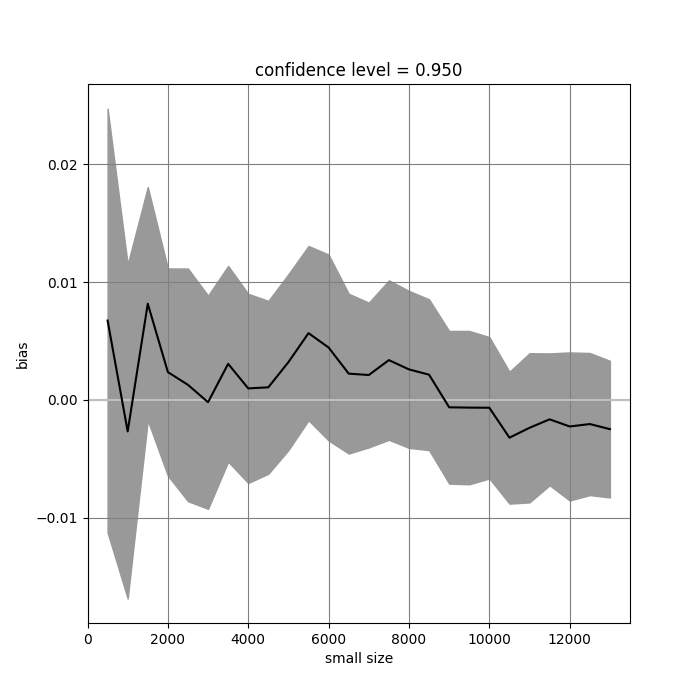}\includegraphics[width=0.32\textwidth,height=0.26\textwidth]{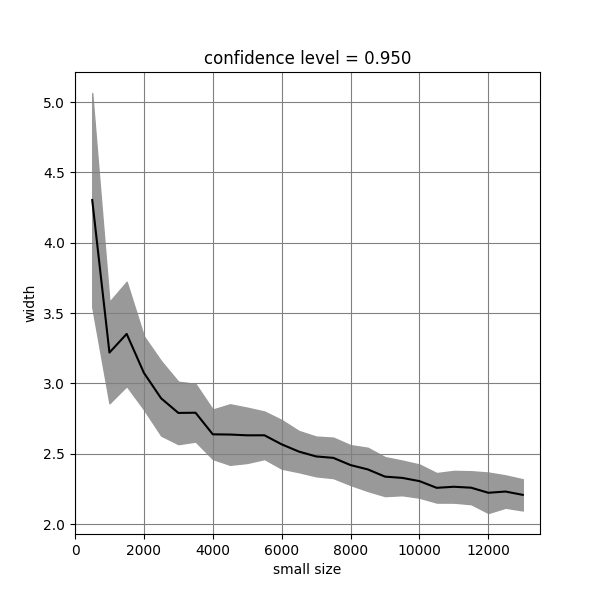}
\label{fig_small_size_index5}}
\hfill
\subfloat[Confidence level = 0.975]{\includegraphics[width=0.32\textwidth,height=0.26\textwidth]{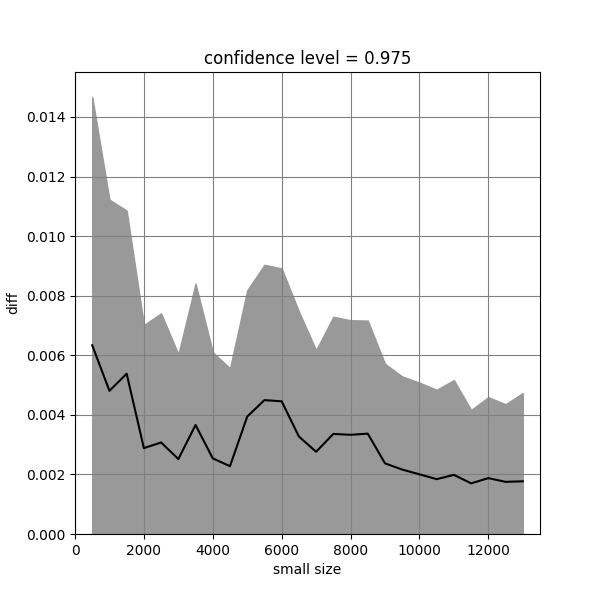}\includegraphics[width=0.32\textwidth,height=0.26\textwidth]{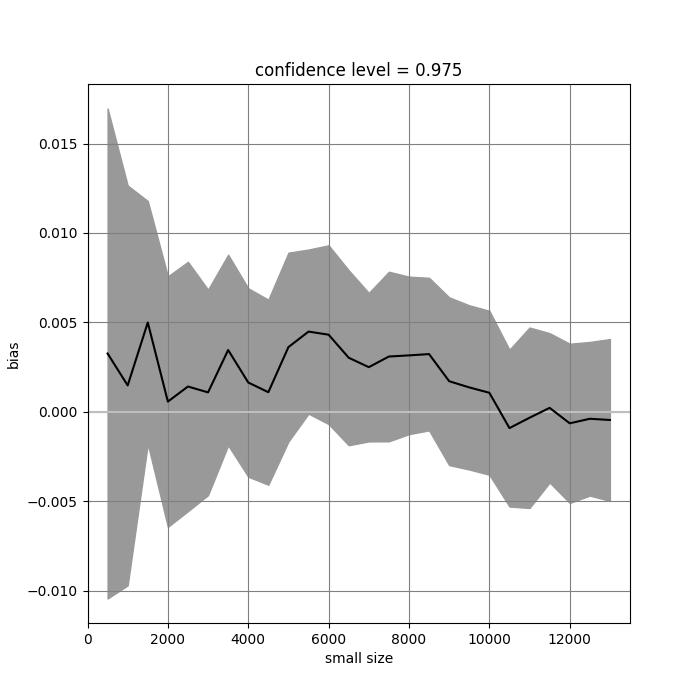}\includegraphics[width=0.32\textwidth,height=0.26\textwidth]{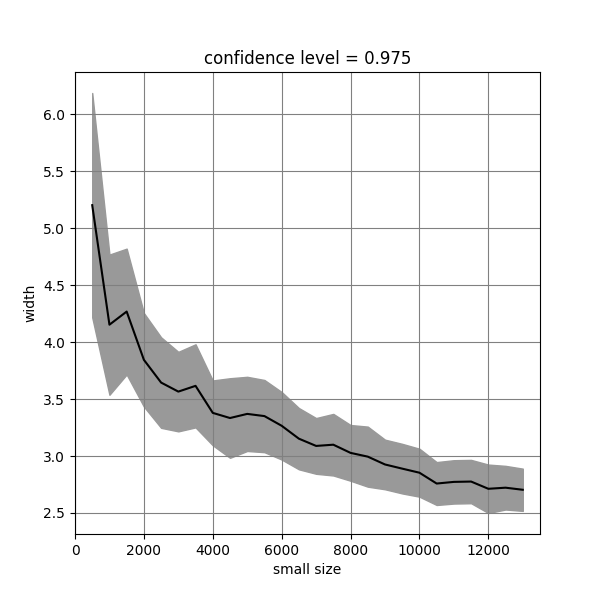}
\label{fig_small_size_index6}}
\hfill
\subfloat[Confidence level = 0.99]{\includegraphics[width=0.32\textwidth,height=0.26\textwidth]{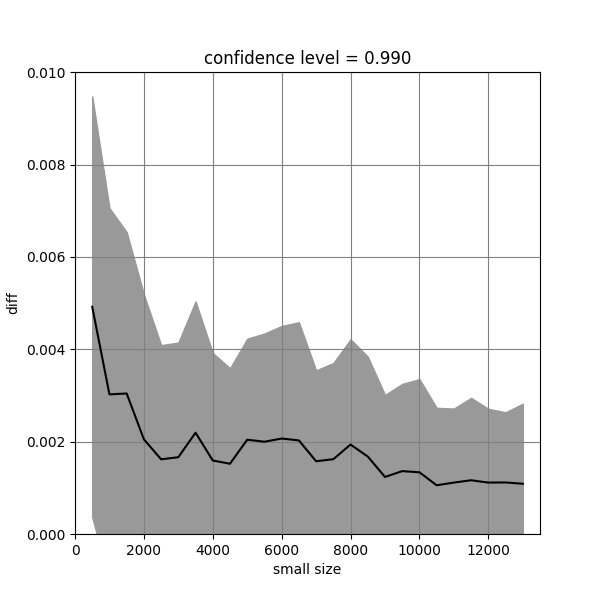}\includegraphics[width=0.32\textwidth,height=0.26\textwidth]{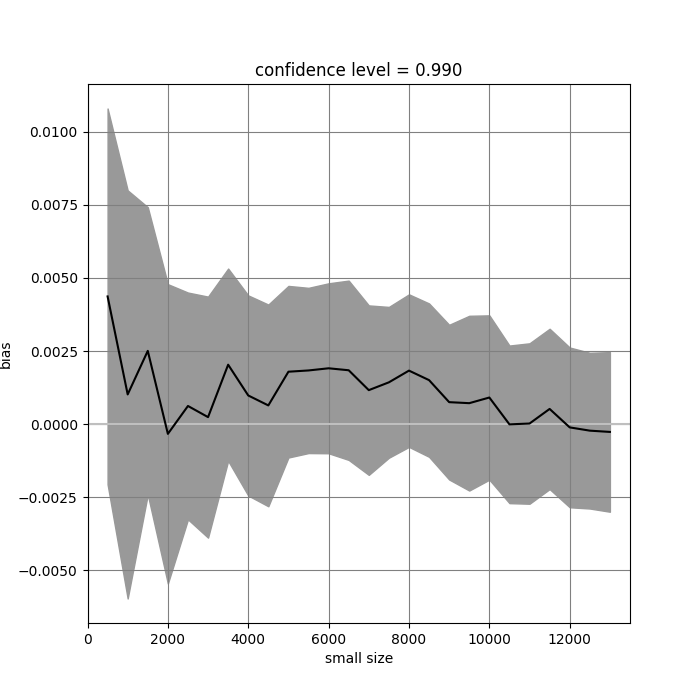}\includegraphics[width=0.32\textwidth,height=0.26\textwidth]{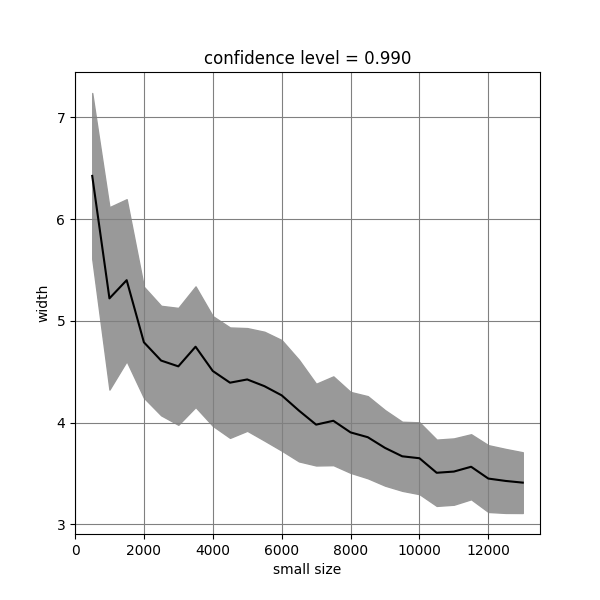}
\label{fig_small_size_index7}}
\hfill
\caption{Experiment B: \emph{diff} (left), \emph{bias} (middle), \emph{width} (right) for different combined calibration and training set size (horizontal axes) and confidence levels, with 95\% confidence intervals.
}
\label{fig:expt_B}
\end{figure}

\begin{figure}[ht]
\centering
\subfloat[Confidence level = 0.2]{\includegraphics[width=0.32\textwidth,height=0.26\textwidth]{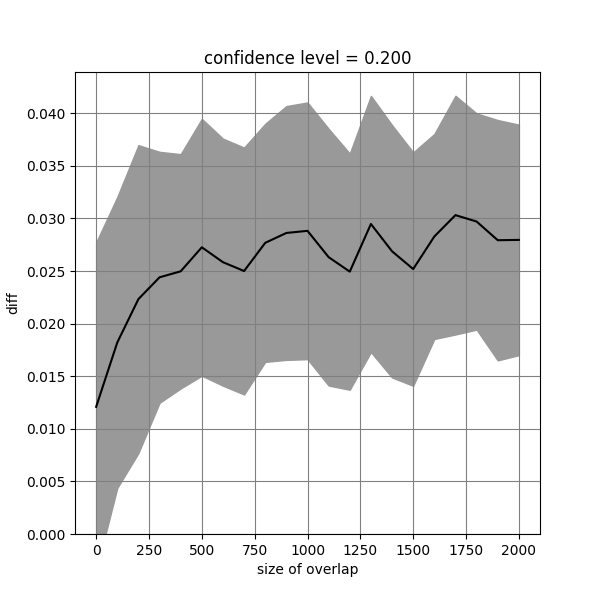}\includegraphics[width=0.32\textwidth,height=0.26\textwidth]{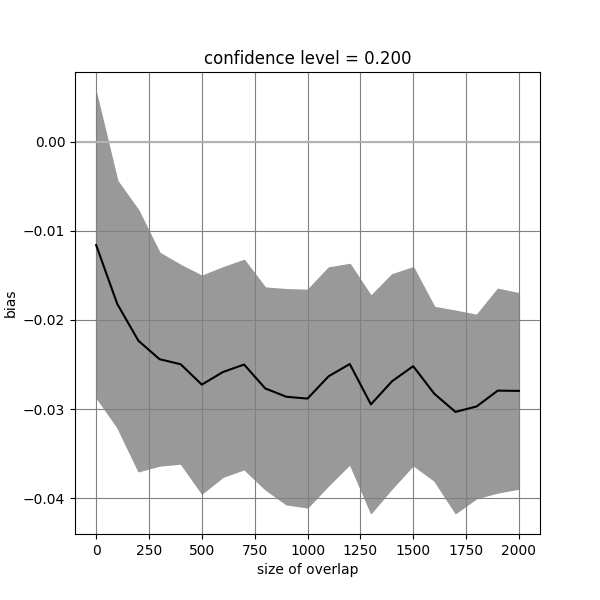}\includegraphics[width=0.32\textwidth,height=0.26\textwidth]{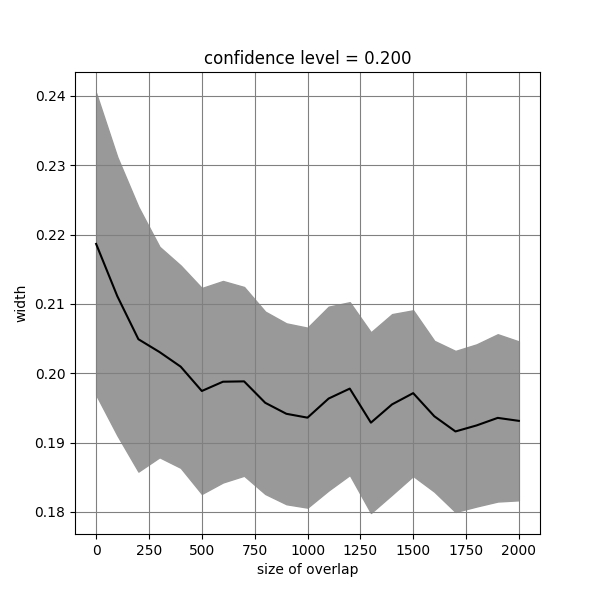}
\label{fig_change_overlap_index0}}
\hfill
\subfloat[Confidence level = 0.4]{\includegraphics[width=0.32\textwidth,height=0.26\textwidth]{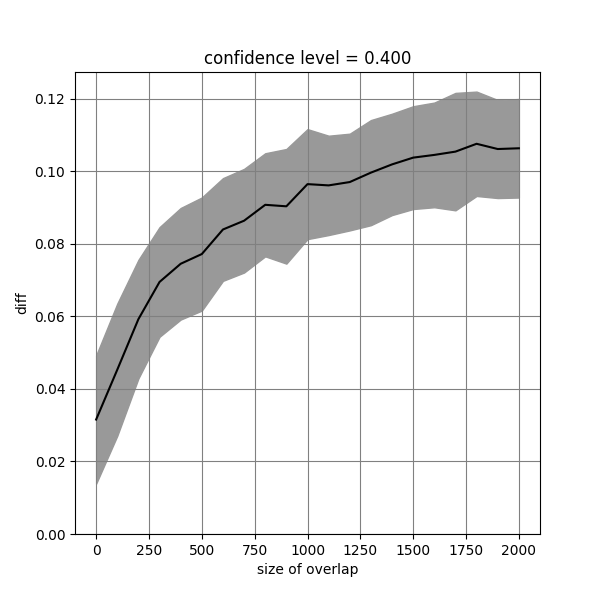}\includegraphics[width=0.32\textwidth,height=0.26 \textwidth]{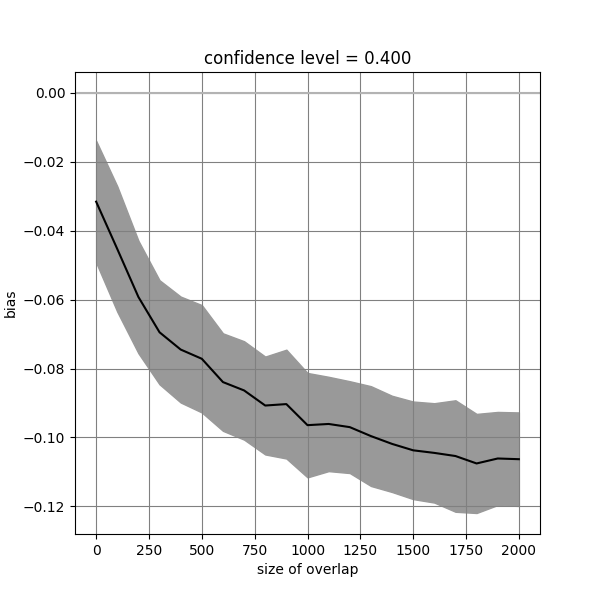}\includegraphics[width=0.32\textwidth,height=0.26\textwidth]{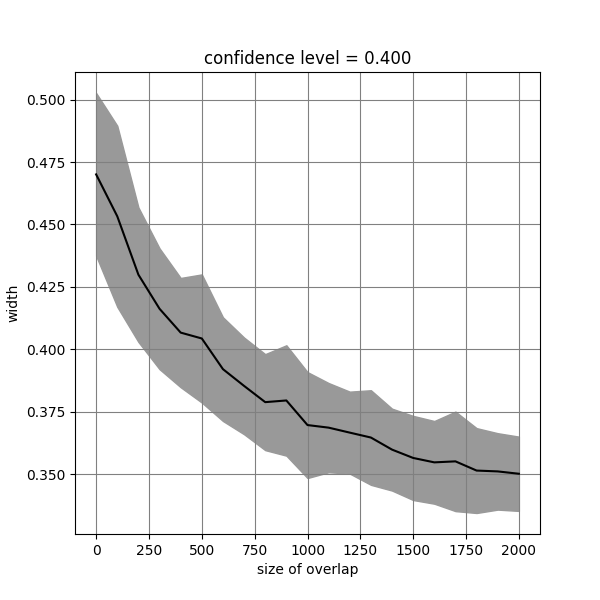}
\label{fig_change_overlap_index1}}
\hfill
\subfloat[Confidence level = 0.6]{\includegraphics[width=0.32\textwidth,height=0.26\textwidth]{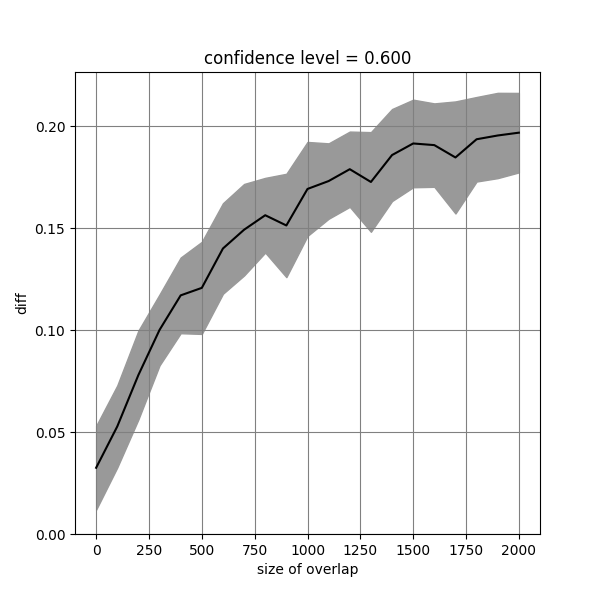}\includegraphics[width=0.32\textwidth,height=0.26\textwidth]{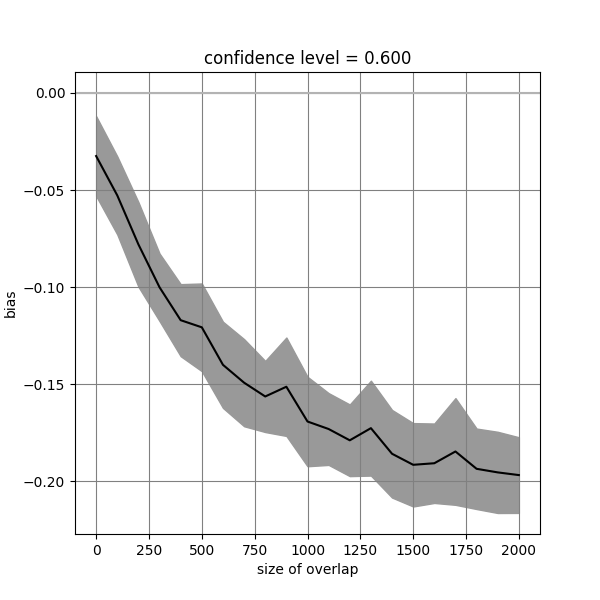}\includegraphics[width=0.32\textwidth,height=0.26\textwidth]{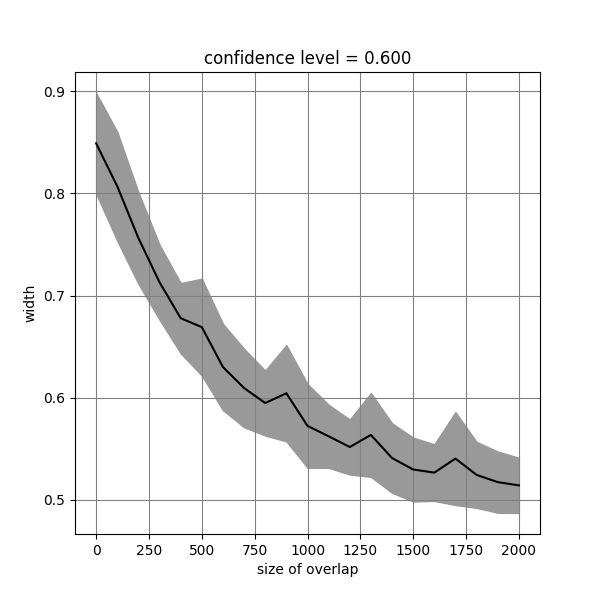}
\label{fig_change_overlap_index2}}
\hfill
\subfloat[Confidence level = 0.8]{\includegraphics[width=0.32\textwidth,height=0.26\textwidth]{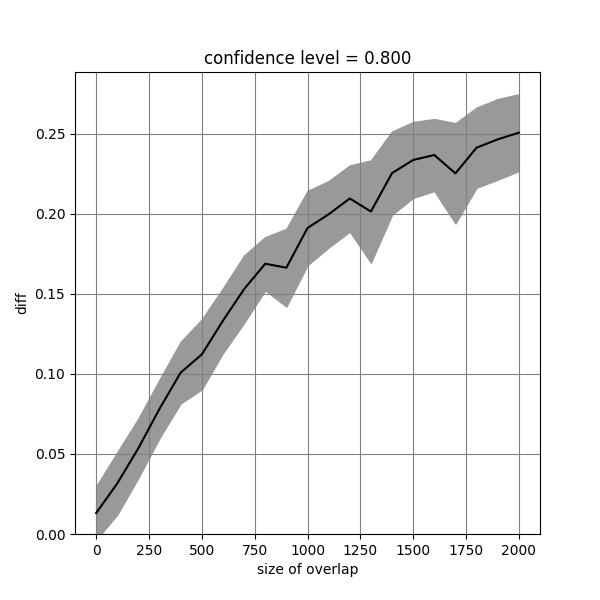}\includegraphics[width=0.32\textwidth,height=0.26\textwidth]{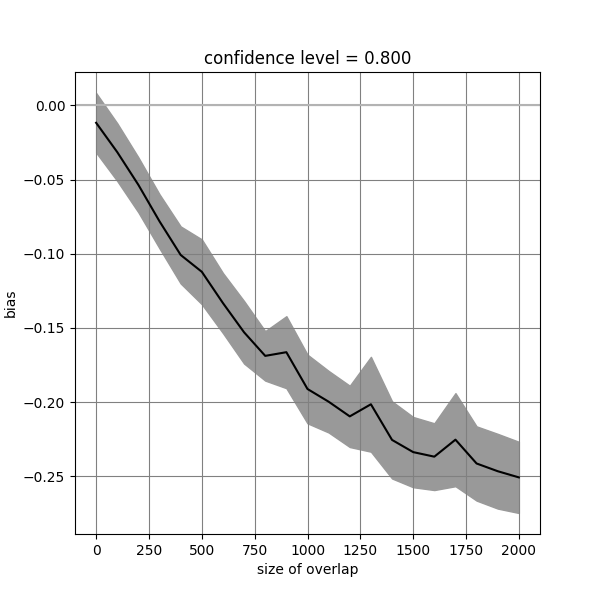}\includegraphics[width=0.32\textwidth,height=0.26\textwidth]{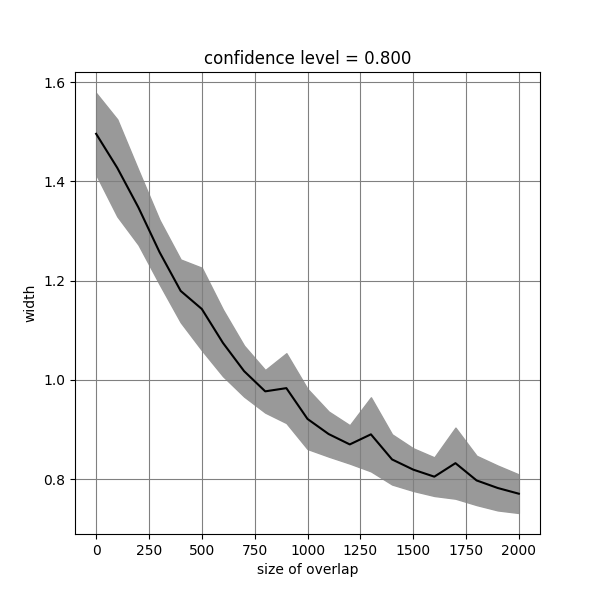}
\label{fig_change_overlap_index3}}
\hfill
\caption{Experiment C: \emph{diff} (left), \emph{bias} (middle), \emph{width} (right) for different training set size (horizontal axes) and confidence levels, with 95\% confidence intervals.
\emph{Continued on next page.}}
\end{figure}
\begin{figure}[ht]
\ContinuedFloat
\subfloat[Confidence level = 0.9]{\includegraphics[width=0.32\textwidth,height=0.26\textwidth]{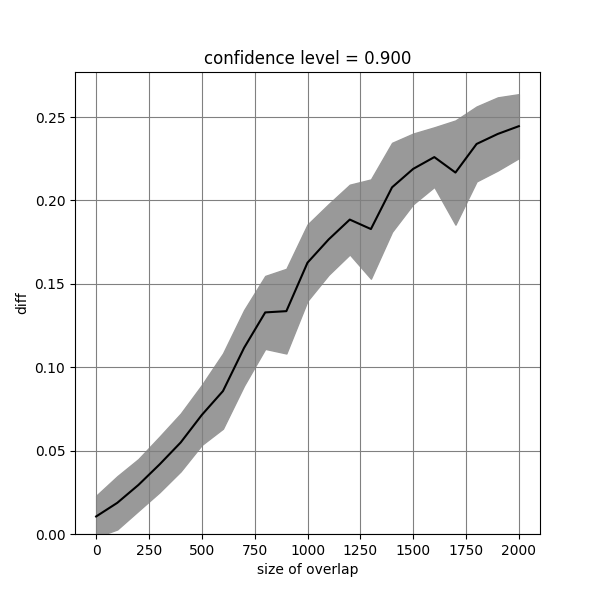}\includegraphics[width=0.32\textwidth,height=0.26\textwidth]{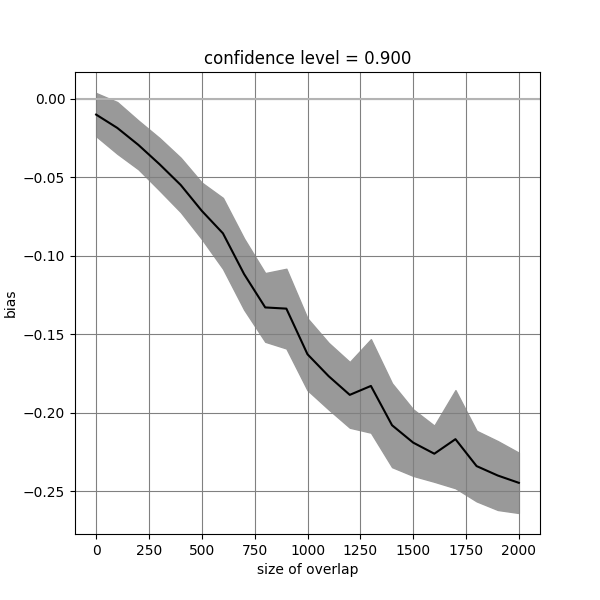}\includegraphics[width=0.32\textwidth,height=0.26\textwidth]{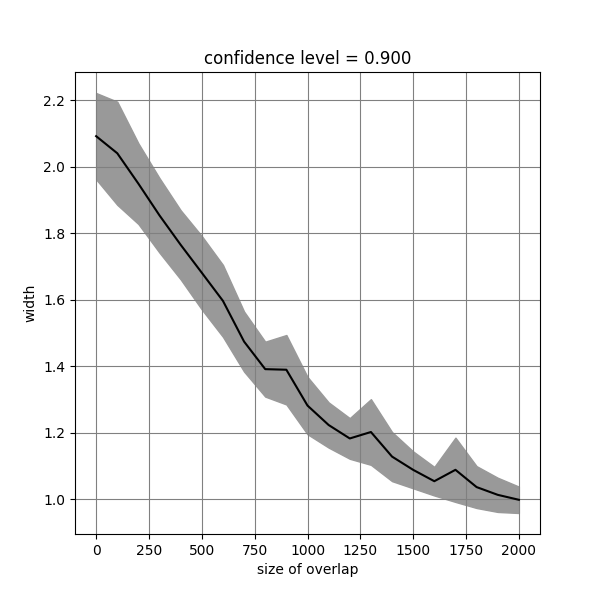}
\label{fig_change_overlap_index4}}
\hfill
\subfloat[Confidence level = 0.95]{\includegraphics[width=0.32\textwidth,height=0.26\textwidth]{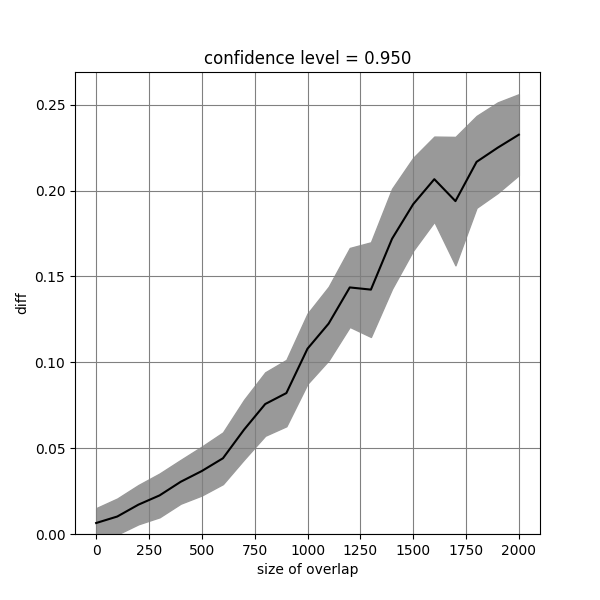}\includegraphics[width=0.32\textwidth,height=0.26\textwidth]{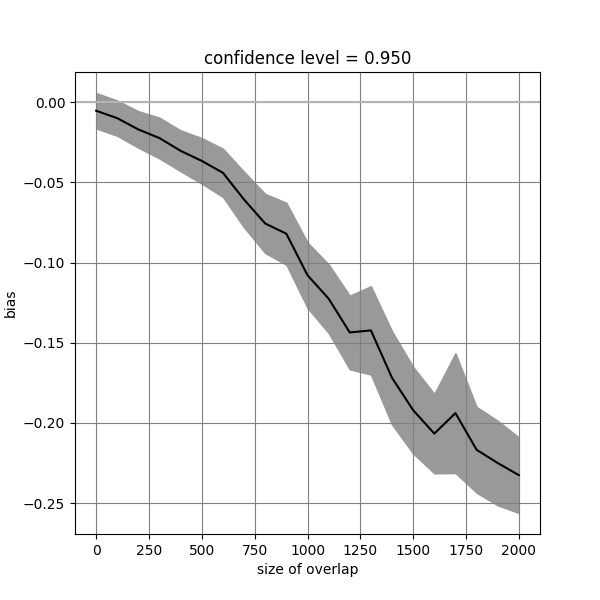}\includegraphics[width=0.32\textwidth,height=0.26\textwidth]{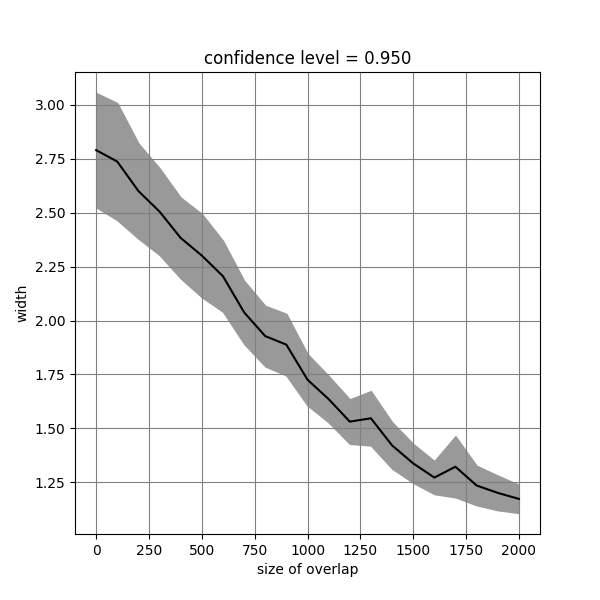}
\label{fig_change_overlap_index5}}
\hfill
\subfloat[Confidence level = 0.975]{\includegraphics[width=0.32\textwidth,height=0.26\textwidth]{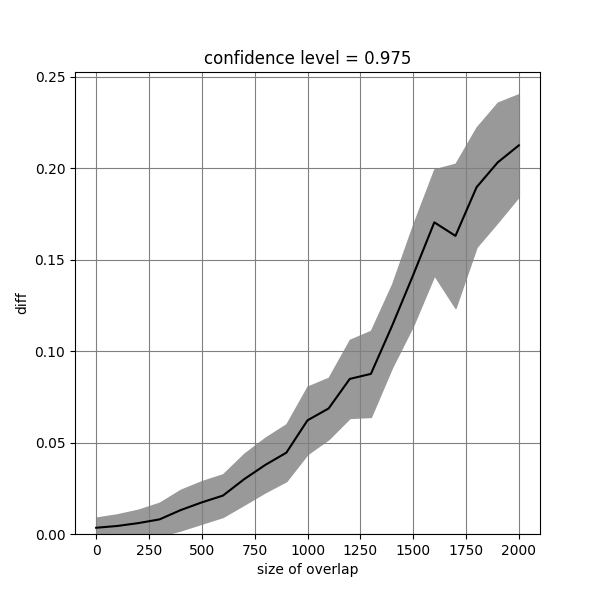}\includegraphics[width=0.32\textwidth,height=0.26\textwidth]{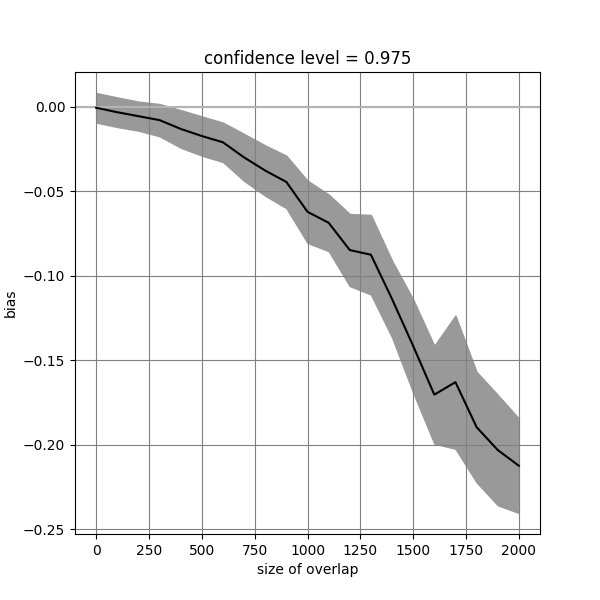}\includegraphics[width=0.32\textwidth,height=0.26\textwidth]{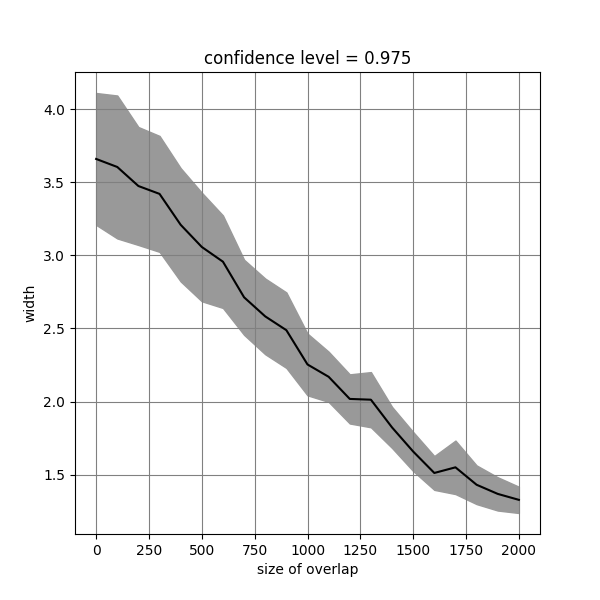}
\label{fig_change_overlap_index6}}
\hfill
\subfloat[Confidence level = 0.99]{\includegraphics[width=0.32\textwidth,height=0.26\textwidth]{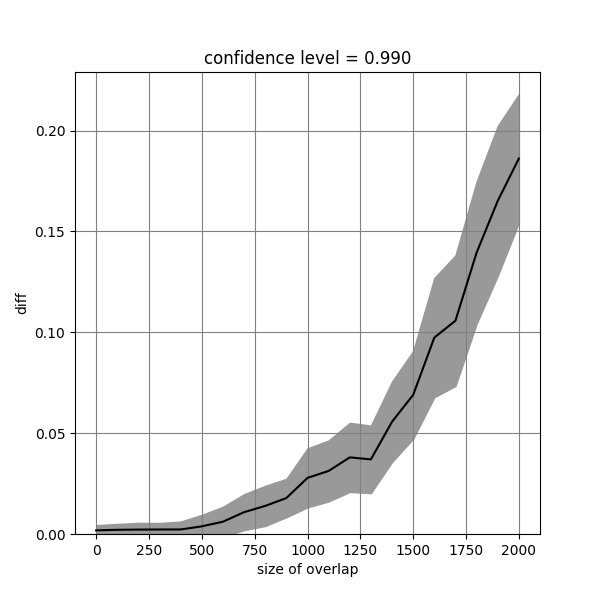}\includegraphics[width=0.32\textwidth,height=0.26\textwidth]{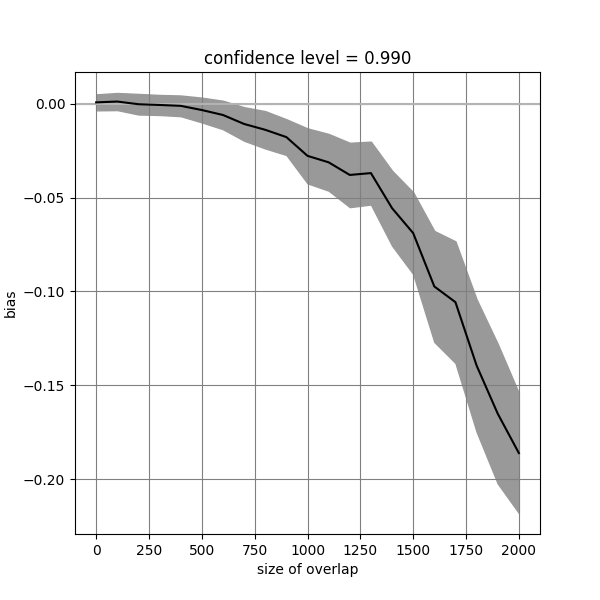}\includegraphics[width=0.32\textwidth,height=0.26\textwidth]{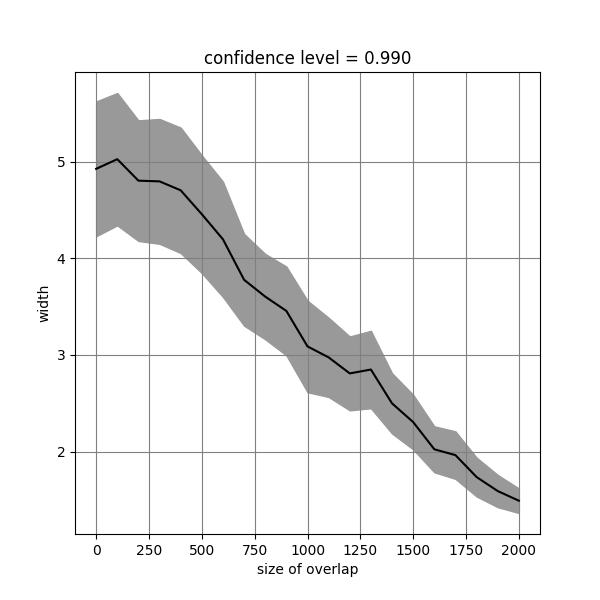}
\label{fig_change_overlap_index7}}
\hfill
\caption{Experiment C: \emph{diff} (left), \emph{bias} (middle), \emph{width} (right) for different training set size (horizontal axes) and confidence levels, with 95\% confidence intervals.}
\label{fig:expt_C}
\end{figure}

\end{document}